\documentclass{article}

\usepackage[main, final]{neurips_2026}
\usepackage[utf8]{inputenc} % allow utf-8 input
\usepackage[T1]{fontenc}    % use 8-bit T1 fonts
\usepackage{hyperref}       % hyperlinks
\usepackage{url}            % simple URL typesetting
\usepackage{booktabs}       % professional-quality tables
\usepackage{amsfonts}       % blackboard math symbols
\usepackage{nicefrac}       % compact symbols for 1/2, etc.
\usepackage{microtype}      % microtypography
\usepackage{xcolor}         % colors
\usepackage{multirow}
\usepackage[table]{xcolor}

\usepackage{amsmath,amsfonts,bm}

\def\eqref#1{equation~\ref{#1}}
\def\1{\bm{1}}

\DeclareMathAlphabet{\mathsfit}{\encodingdefault}{\sfdefault}{m}{sl}
\SetMathAlphabet{\mathsfit}{bold}{\encodingdefault}{\sfdefault}{bx}{n}

\newcommand{\hidden}[1]{}
\usepackage{url}
\usepackage{graphicx}
\usepackage[most]{tcolorbox}

\usepackage{amsmath}
\usepackage{amssymb}
\usepackage{mathtools}
\usepackage{amsthm}

\usepackage{subcaption}

\usepackage[capitalize,noabbrev]{cleveref}

\theoremstyle{plain}
\newtheorem{theorem}{Theorem}
\newtheorem{lemma}[theorem]{Lemma}
\newtheorem{corollary}[theorem]{Corollary}
\theoremstyle{definition}

\theoremstyle{remark}

\usepackage[textsize=tiny]{todonotes}

\title{Why Transformers Struggle with Distribution-Independent In-Context Learning}

\author{%
  Omar Naim \\
  Universit\'e de Toulouse \\
 \small{\texttt{omar.naim.docs@gmail.com}}
  \And
  J\'er\^ome Bolte \\
  Toulouse School of Economics \\
  \small{\texttt{jbolte@ut-capitole.fr}}
  \And
  Nicholas Asher \\
  Universit\'e de Toulouse \\
 \small{\texttt{nicholas.asher@irit.fr}}
}

\begin{document}

\maketitle

\begin{abstract}
With in-context learning (ICL), transformer models adapt to new tasks from illustrative examples at inference time without updating their parameters. A common interpretation is that transformers implement implicit learning algorithms in their forward pass. We test this interpretation and ICL capability under distribution shift in noiseless polynomial regression, where the context uniquely determines the target function. Models that are near-perfect in distribution exhibit sharp performance degradation under shifts in the input or target-function distribution. This behavior persists across polynomial degrees and multiple training interventions, and in higher-dimensional nonlinear regression. We identify two architectural mechanisms that explain this breakdown: output saturation at finite boundary values linked to final normalization, and attention concentration on a small subset of the context, which can limit context-dependent adaptation required for ICL. Controlled experiments, architectural ablations, targeted interventions, and theoretical analysis provide strong evidence for both mechanisms and their consequence: strong in-distribution ICL does not imply distribution-independent ICL, which calls for caution when using ICL instead of explicit retraining or adaptation under distribution shift.
\end{abstract}

\section{Introduction}

Transformer models can solve new tasks from a few input--output examples provided at inference time, without any parameter updates. Known as in-context learning (ICL), this ability makes transformer models highly flexible and attractive~\citep{brown:etal:2020}. A natural and influential interpretation or ``received view'' is that transformers perform ICL by implicitly implementing a learning algorithm in their forward pass, processing examples to approximate procedures such as gradient descent or least-squares regression~\citep{von-oswald:etal:2022,akyurek:etal:2022,dai2023can}.  

We show that the assumption behind ICL---that models can find the appropriate solution from context alone---can fail sharply. In controlled regression tasks where the ground-truth solution is known and where exact extrapolation is possible for classical estimators, transformers exhibit a qualitatively distinct failure mode under distribution shift. Performance remains near-perfect in-distribution, but degrades abruptly once the test distribution is moderately shifted. This ``cliff-shaped'' collapse contrasts with least-squares and kernel methods, which degrade smoothly or remain exact when the model class is correctly specified. The phenomenon persists across polynomial degrees and a wide range of training interventions.

This behavior challenges the received view. By construction, transformers can simulate gradient-based updates~\citep{von-oswald:etal:2022,mahankali2023one}. But if a model were applying such an algorithm, performance would be governed primarily by the conditioning of the problem and by model-class mismatch. It should not exhibit an abrupt, architecture-specific collapse under moderate scale shifts. The observed failure instead points to structural constraints of the transformer architecture.

We identify two complementary architectural bottlenecks.

\textbf{(i) A readout-side scale bottleneck induced by final normalization.}
Final normalization bounds the readout representation, causing predictions to saturate at fixed boundary values. Removing it eliminates this hard bound, but does not restore reliable ICL under distribution shift.

\textbf{(ii) A context bottleneck induced by attention.}
Attention is the mechanism by which the query token aggregates the in-context examples. Under distribution shift, however, attention weights can concentrate on a small part of the prompt, limiting the effective context available to the query update precisely when ICL requires context-dependent adaptation.

Together, these mechanisms help explain why standard transformer architectures do not, in general, achieve distribution-independent in-context learning, even in settings where exact generalization is theoretically possible. This distinction is not merely theoretical: transformer-based ICL is increasingly used as a substitute for explicit retraining or task-specific adaptation, including in tabular learning~\citep{tabpfn2022, contexttab2025}, reinforcement learning~\citep{wu2025mixture}, and scientific applications~\citep{seifner2025context}.
These systems are attractive because they use relatively small models that can be trained efficiently and deployed as inference-time adapters without parameter updates. However, real-data deployments may involve distribution shifts that are not revealed by matched-distribution evaluation. Understanding the mechanisms and limits of ICL generalization is therefore both scientifically and practically urgent.

\section{Problem Setup and Experimental Framework}
 
\subsection{In-context learning.}
We use polynomial regression as a controlled testbed for studying in-context learning under distribution shift. This setting has two useful properties. First, the ground-truth solution is known: once sufficiently many input--output examples are provided, the target polynomial is uniquely determined within its function class, and classical polynomial regression can recover it exactly. Second, the training and test distributions can be varied independently while keeping the underlying function class fixed. This lets us distinguish matched-distribution ICL from out-of-distribution ICL without changing the target task itself.

Concretely, given a prompt

\[
  (x_1, g(x_1), \ldots, x_p, g(x_p),\, x),
\]

the model must predict $g(x)$ using only the input--output examples provided in the same sequence. We consider polynomials of degree at most $n$, denoted $\mathbb{R}_n[X]$, for $n \in \{1,\ldots,6\}$.

Let $\mathcal{D}_{\mathcal I}$ denote the input distribution and $\mathcal{D}_{\mathcal F}$ the distribution over polynomial coefficients. During training, inputs are sampled as $x \sim \mathcal{D}_{\mathcal I}$, and functions $g$ are generated by sampling their coefficients independently from $\mathcal{D}_{\mathcal F}$. In our experiments, both $\mathcal{D}_{\mathcal I}$ and $\mathcal{D}_{\mathcal F}$ are uniform distributions on $[-1,1]$.

For each degree $n$, we train a transformer model from scratch on this distribution. We then test the model on sequences with inputs drawn from $\mathcal{D}^{\text{test}}_{\mathcal I}$ and functions with coefficients drawn from $\mathcal{D}^{\text{test}}_{\mathcal F}$.

\paragraph{Training.}
We train transformers using the standard supervised ICL objective of \citet{garg:etal:2022}. Each training sequence is generated by sampling a function $g \sim \mathcal{D}_{\mathcal F}$ and evaluating it at input points $x_1,\ldots,x_{k+1} \sim \mathcal{D}_{\mathcal I}$. The model is then trained autoregressively to predict each next function value from the preceding input--output context together with the query input.

\begin{equation}
\label{eq:autoregressive}
\min_{\theta} \;
\mathbb{E}_{g \sim \mathcal{D}_\mathcal{F}}
\left[
  \mathbb{E}_{x_1, \dots, x_{k+1} \sim \mathcal{D}_\mathcal{I}}
  \sum_{i=1}^{k}
  \ell\!\left(
    g(x_{i+1}),\,
    f^{\theta}(x_1, g(x_1), \dots, x_{i+1})
  \right)
\right],
\end{equation}
where $\ell(\cdot,\cdot)$ is the squared error loss. Details of the architecture, training procedure, and evaluation protocol are given in Appendix~\ref{sec:trainingdetails}.\footnote{Code:
\url{https://github.com/omyokun/icl-generalization}}

\subsection{Two Notions of In-Context Learning}
\label{sec:notions}
 
We distinguish two notions of ICL, which correspond to qualitatively different generalization requirements. Throughout, $\epsilon > 0$ denotes a fixed tolerance, and $p_0$ denotes the number of examples needed to determine a function within $\mathcal{F}$ ($p_0 = n+1$ for $\mathbb{R}_n[X]$).

\noindent\textbf{Definition (In-distribution ICL, ICL$_1$).} A transformer $\hat{f}_\theta$ exhibits in-distribution ICL for a function class $\mathcal{F}$ if, for any $g \sim \mathcal{D}_\mathcal{F}$, any $p \geq p_0$, any in-context examples $\{(x_i,g(x_i))\}_{i=1}^p$ with $x_i \sim \mathcal{D}_I$, and any query point $x \sim \mathcal{D}_I$, the prediction satisfies
\[
  \|\hat{f}_\theta(x_1,g(x_1),\ldots,x_p,g(x_p),x) - g(x)\|_2 < \epsilon.
\]

\noindent\textbf{Definition (Distribution-independent ICL, ICL$_2$).} A transformer exhibits distribution-independent ICL for $\mathcal{F}$ if it can accurately approximate any $g \in \mathcal{F}$ for arbitrary query points $x \in \mathrm{Dom}(g)$ (the full domain on which $g$ is defined), regardless of input distribution. Formally, for all $g \in \mathcal{F}$, all $x \in \mathrm{Dom}(g)$, and all choices of in-context examples $\{(x_i, g(x_i))\}_{i=1}^p$ that uniquely determine $g$ within $\mathcal{F}$,
\[
  \|\hat{f}_\theta(x_1,g(x_1),\ldots,x_p,g(x_p),x) - g(x)\|_2 < \epsilon.
\]

$\mathrm{ICL}_2$ captures the behavior expected of a model that has genuinely learned an analytic method for computing functions in $\mathcal{F}$: ordinary least squares (OLS) satisfies $\mathrm{ICL}_2$ for linear functions; polynomial regression satisfies $\mathrm{ICL}_2$ for polynomial functions. $\mathrm{ICL}_1$ is the weaker notion prevalent in the literature. Our central finding is that transformers reliably achieve $\mathrm{ICL}_1$ but systematically fail $\mathrm{ICL}_2$.

\section{$\mathrm{ICL}_1$ Does Not Imply $\mathrm{ICL}_2$}
\label{sec:failure}

\begin{figure}[!h]
\center
\includegraphics[width=\linewidth]{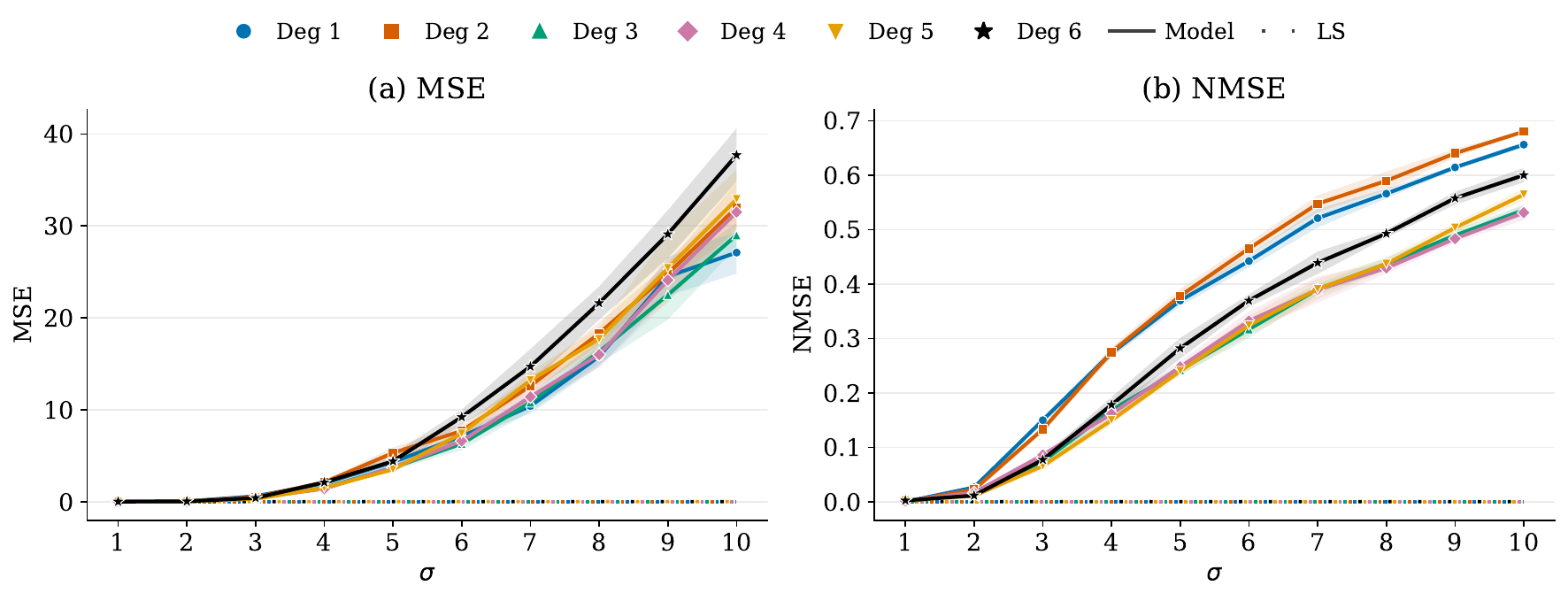}
\caption{Mean squared error (MSE; left) and normalized MSE (NMSE; right) under distribution shift. Each Transformer (12 layers, 8 attention heads, $d_{\mathrm{emb}}=256$) is trained and evaluated on a single polynomial degree from 1 to 6. Training inputs and coefficients are sampled from $\mathcal{U}(-1,1)$. At test time, inputs follow the same distribution, while coefficients are sampled from $\mathcal{U}(-\sigma,\sigma)$. Solid curves show Transformer results, and dotted curves show least-squares baselines. Results are averaged over five evaluation seeds; shaded regions indicate $\pm$ one standard deviation. The evaluation protocol is described in  Appendix~\ref{sec:trainingdetails}.}
\label{deg}
\end{figure}

\paragraph{Matched-distribution ICL.}
We first verify that the trained transformers exhibit strong in-distribution in-context learning. When the training and test distributions match, models trained on polynomial regression tasks achieve near-zero error across degrees $1$ through $6$ (Figure~\ref{deg} and Table~\ref{table:3a}, $\sigma=1$), comparable to OLS when the polynomial degree is correctly specified. This confirms that the models solve the standard matched-distribution ICL task.

Such behavior is often interpreted as evidence that transformers implement an implicit learning algorithm at inference time, such as gradient descent or regression \citep{von-oswald:etal:2022, akyurek:etal:2022, dai2023can}. This interpretation is plausible in principle: for example, a single-layer transformer with linear attention can be constructed to implement one gradient-descent step for linear regression (Appendix~\ref{sec:theory}). However, matched-distribution success alone does not determine whether the learned procedure generalizes distribution-independently.

\paragraph{A cliff under distribution shift.}
We therefore test the same models under controlled distribution shifts. During training, inputs and coefficients are sampled from $\mathcal{U}(-1,1)$. At test time, we expand the coefficient distribution to $\mathcal{U}(-\sigma,\sigma)$ while keeping the input distribution fixed. Classical polynomial regression remains exact under this shift, since the model class is correctly specified and the in-context examples uniquely determine the target polynomial.

Transformers show a qualitative difference. Their error remains negligible near the training scale, but increases abruptly by orders of magnitude once $\sigma$ reaches a moderate value (Figure~\ref{deg}). This degradation is not an artifact of the raw MSE increasing with target magnitude: the normalized MSE (NMSE) also grows substantially %under coefficient scaling 
(Figure \ref{deg}). We call this behavior a \emph{cliff-shaped collapse}. In contrast, OLS remains exact across the same range of shifts.

The collapse is robust across polynomial degrees $1$ through $6$. Although the underlying regression problems vary in degree, the transformer error curves are strikingly similar across tasks. This is not the behavior expected from a standard regression algorithm, whose error under shift should be governed primarily by identifiability, conditioning, and model-class mismatch.

\newpage

\paragraph{Comparison with kernel baselines.}
An alternative is that the transformer behaves like an implicit kernel method \citep{han2025understanding,shen2025understanding}. We therefore compare against Nadaraya--Watson regression and kernel ridge regression with multiple kernels, using the same in-context examples at test time (Table \ref{tab:kernel}). These kernel methods degrade smoothly under scale shift, while the transformer's abrupt, cliff-shaped collapse becomes orders of magnitude worse under moderate shifts. Thus, the assumption that the transformer executes one of these kernel estimators does not explain the observed failure.

\begin{table}[!ht]
  \centering
  \resizebox{\linewidth}{!}{
  \begin{tabular}{lcccccc}
    \toprule
    & $\sigma{=}1$ & $\sigma{=}2$ & $\sigma{=}3$ & $\sigma{=}5$
    & $\sigma{=}7$ & $\sigma{=}10$ \\
    \midrule
    Transformer  & $9{\times}10^{-6}$ & $0.04$ & $0.62$ & $4.29$ & $10.36$ & $27.06$ \\
    NW (RBF)     & $2.3{\times}10^{-2}$ & $0.09$ & $0.27$ & $0.74$ & $1.14$ & $2.75$ \\
    KRR (RBF)    & $0.002$ & $0.008$ & $0.022$ & $0.057$ & $0.095$ & $0.230$ \\
    KRR (Poly)   & $2.2{\times}10^{-5}$ & $9.4{\times}10^{-5}$ & $2.1{\times}10^{-4}$
                 & $6.7{\times}10^{-4}$ & $1.1{\times}10^{-3}$ & $2.3{\times}10^{-3}$ \\
    \bottomrule
  \end{tabular}
  }
  \caption{The evolution of MSE for degree-1 polynomial functions distribution shift, comparing the Transformer with Nadaraya--Watson regression (NW) and kernel ridge regression (KRR). Test inputs are sampled from $\mathcal{U}(-1,1)$ and coefficients from $\mathcal{U}(-\sigma,\sigma)$. All methods receive the same in-context examples. Kernel hyperparameters are specified in Appendix~\ref{sec:trainingdetails}.}
  \label{tab:kernel}
\end{table}

%\vspace{-0.1cm}

\paragraph{Training and representation interventions do not remove the failure.} We also test whether the collapse is an artifact of the training setup or input--output representation. We consider structured supervision, longer training, expanded training support, curriculum strategies, and alternative encodings. We also tested normalizing each prompt by its own label scale and rescaling the prediction back afterward, to control for labels leaving the range seen during training. These interventions change the quantitative error levels but do not remove the qualitative failure pattern (Appendix~\ref{sec:rulingout}). This suggests that the phenomenon is not resolved by the interventions considered here, and motivates an architectural analysis.

\begin{figure*}[!ht]
    \centering
    \includegraphics[width=\textwidth]{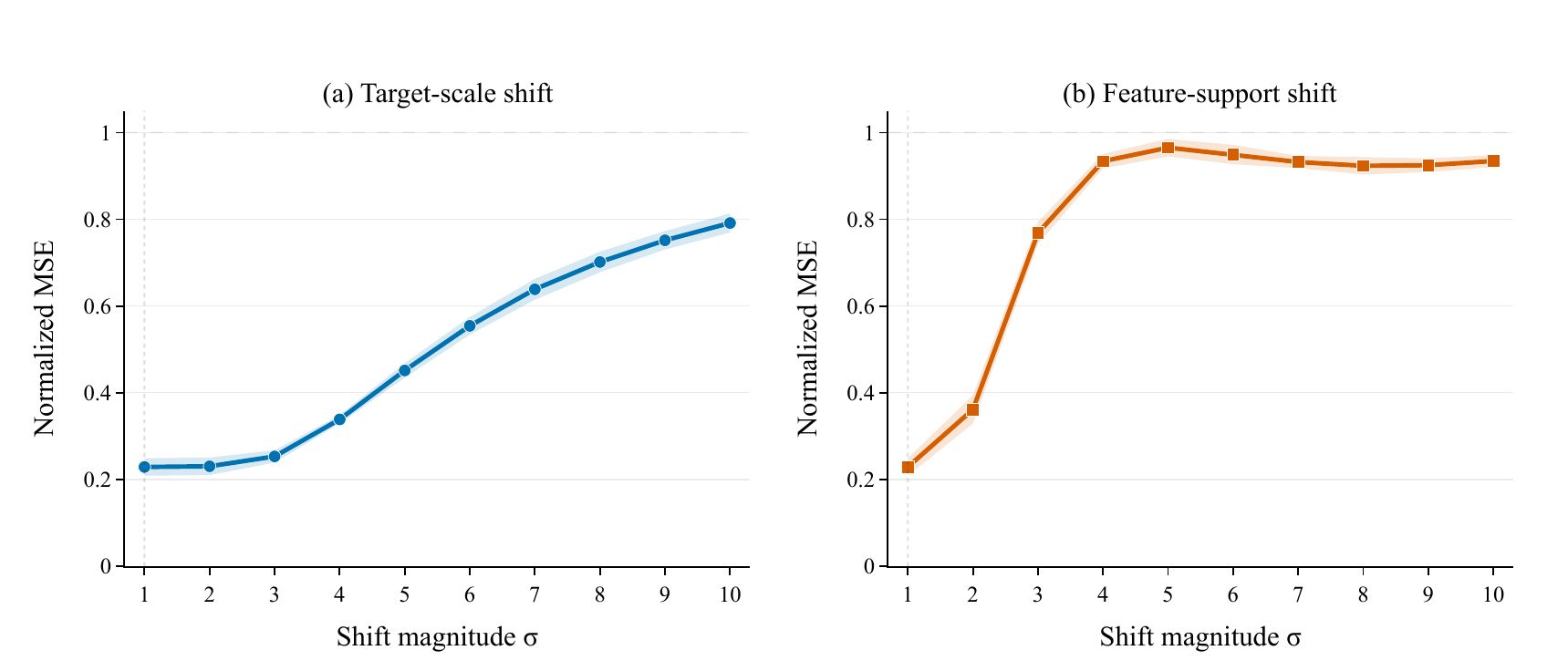}
    \caption{
    NMSE under distribution shift in the 10-dimensional nonlinear regression setting described in Appendix \ref{app:tabular}. (a) Target-scale shift, where the target output is scaled while the inputs remain sampled from $\mathcal{U}(-1,1)^{10}$ . (b) Feature-support shift, where the target function is unchanged and the input support is expanded from $\mathcal{U}(-1,1)^{10}$ to $\mathcal{U}(-\sigma,\sigma)^{10}$. Curves show the mean across five independently generated test draws using a fixed trained model; shaded regions denote $\pm$ standard deviation. The horizontal dashed line at $\mathrm{NMSE}=1$ corresponds to the zero-prediction baseline.
    }
    \label{fig:tabular_shift}
\end{figure*}

\paragraph{Beyond scalar polynomial regression.} To test whether the observed failure is specific to one-dimensional polynomial tasks, we consider a higher-dimensional nonlinear ICL regression setting inspired by the synthetic-prior paradigm of  TabPFN \citep{tabpfn2022} with $10$-dimensional inputs (see Appendix~\ref{app:tabular} for full details). The same qualitative degradation appears under distribution shift, showing that the observed OOD failure is not restricted to scalar polynomial regression (Figure \ref{fig:tabular_shift}).

\newpage

\paragraph{Summary.}
These experiments separate matched-distribution ICL from distribution-independent ICL. Transformers can achieve strong $\mathrm{ICL}_1$ while failing $\mathrm{ICL}_2$ in settings where classical estimators generalize exactly. The sharpness and robustness of the collapse suggest that the failure is not merely a consequence of statistical difficulty, but is tied to architectural mechanisms analyzed in the following sections.

\section{A Readout-Side  Bottleneck}
\label{boundaryvalues}
To understand the architectural origin of this failure, we first examine how the model's predictions behave as they move beyond the training distribution. \citet{naim:asher:2024b} observed for linear functions that predictions eventually stop tracking the target and saturate at finite limits, which they termed \emph{boundary values}. We find the same behavior for polynomial regression across degrees (Figure~\ref{sequence1}). Once the target moves beyond this output range, the prediction error necessarily grows, providing a direct connection between boundary-value saturation and the cliff-shaped degradation observed in Figure~\ref{deg}. The same behavior also appears in the higher-dimensional nonlinear setting (Appendix~\ref{app:tabular}). We next ask what in the Transformer architecture gives rise to this behavior.

\begin{figure}[!ht]
\centering
\includegraphics[width=4.6cm]{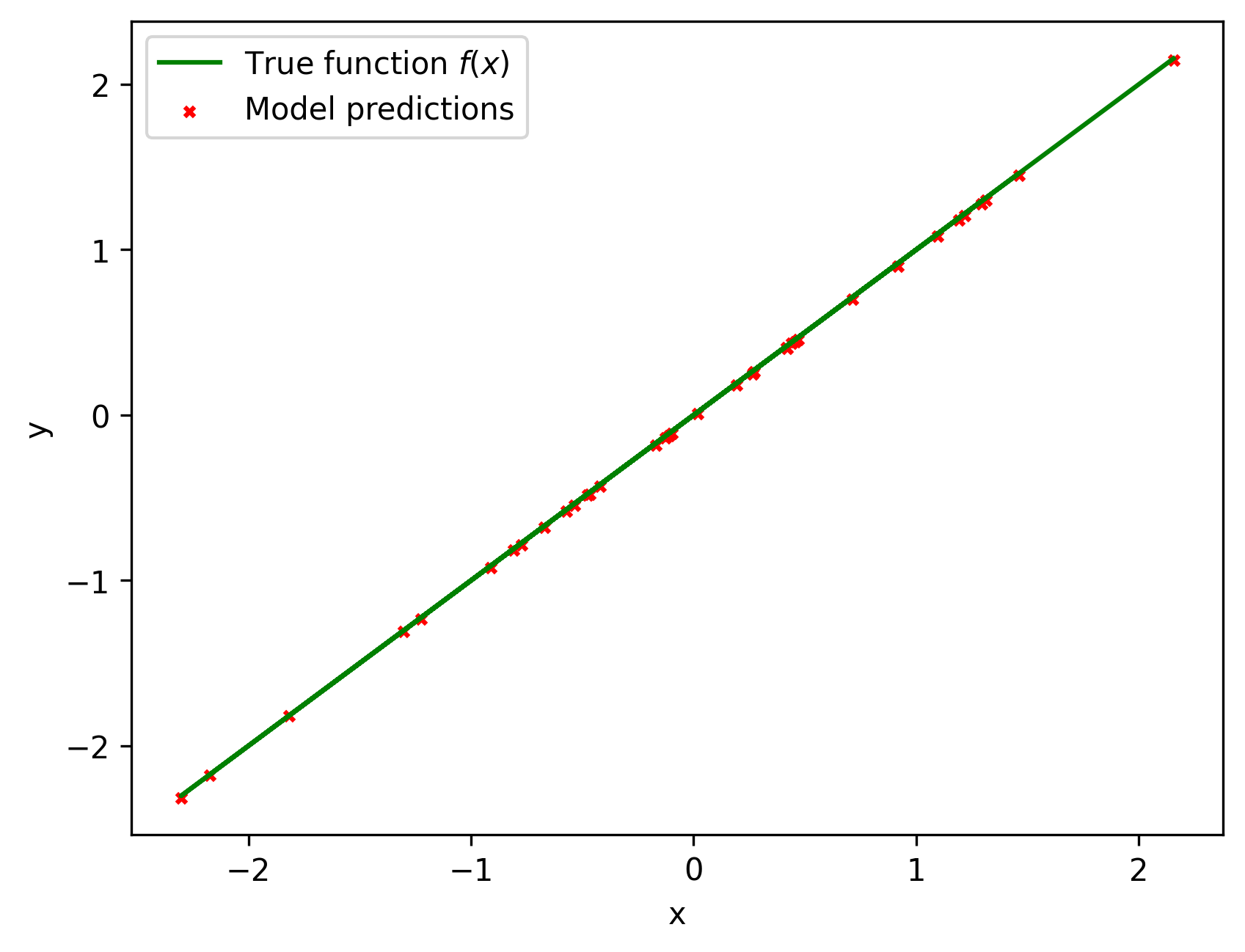}
\includegraphics[width=4.6cm]{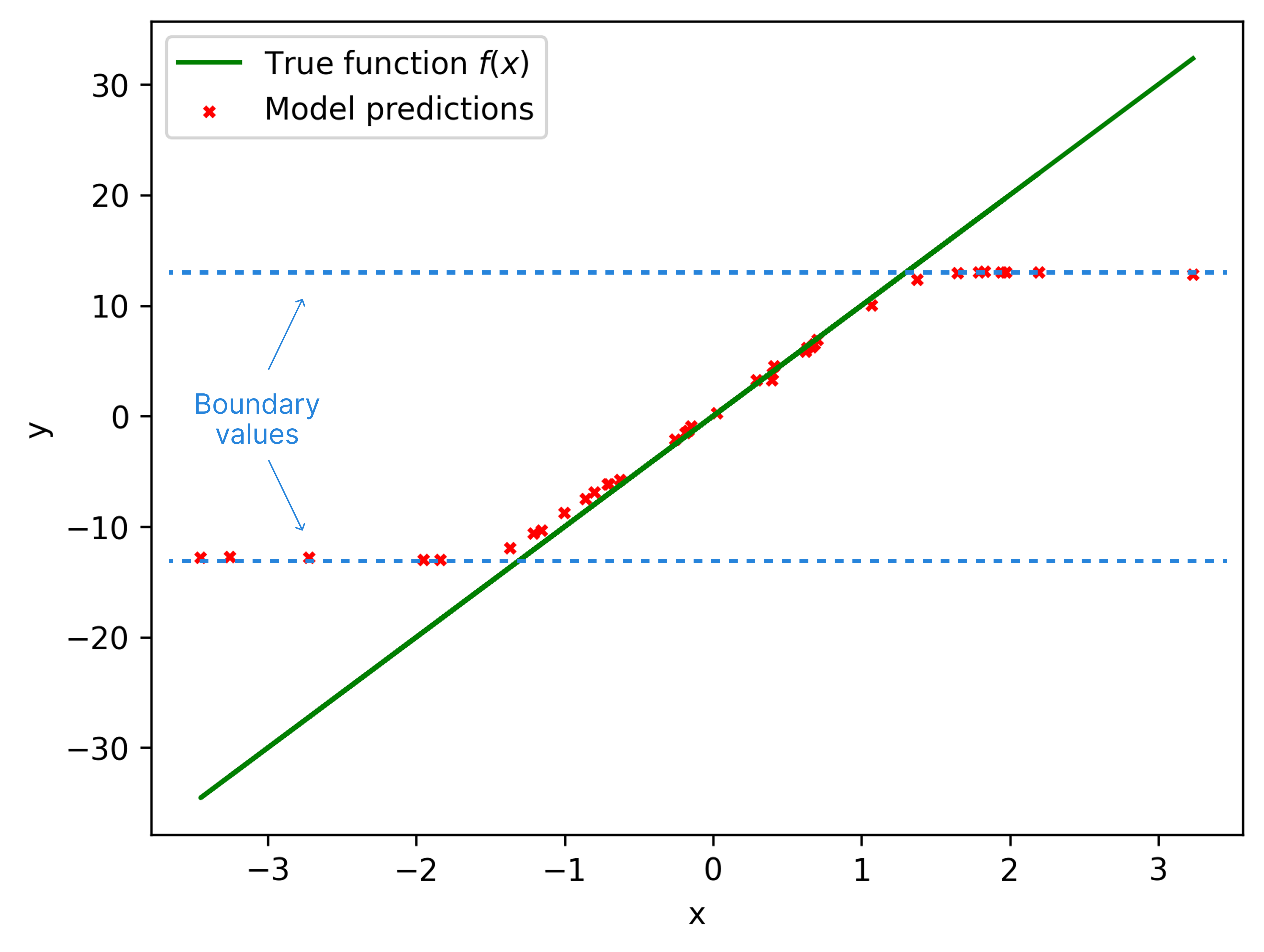}
\includegraphics[width=4.6cm]{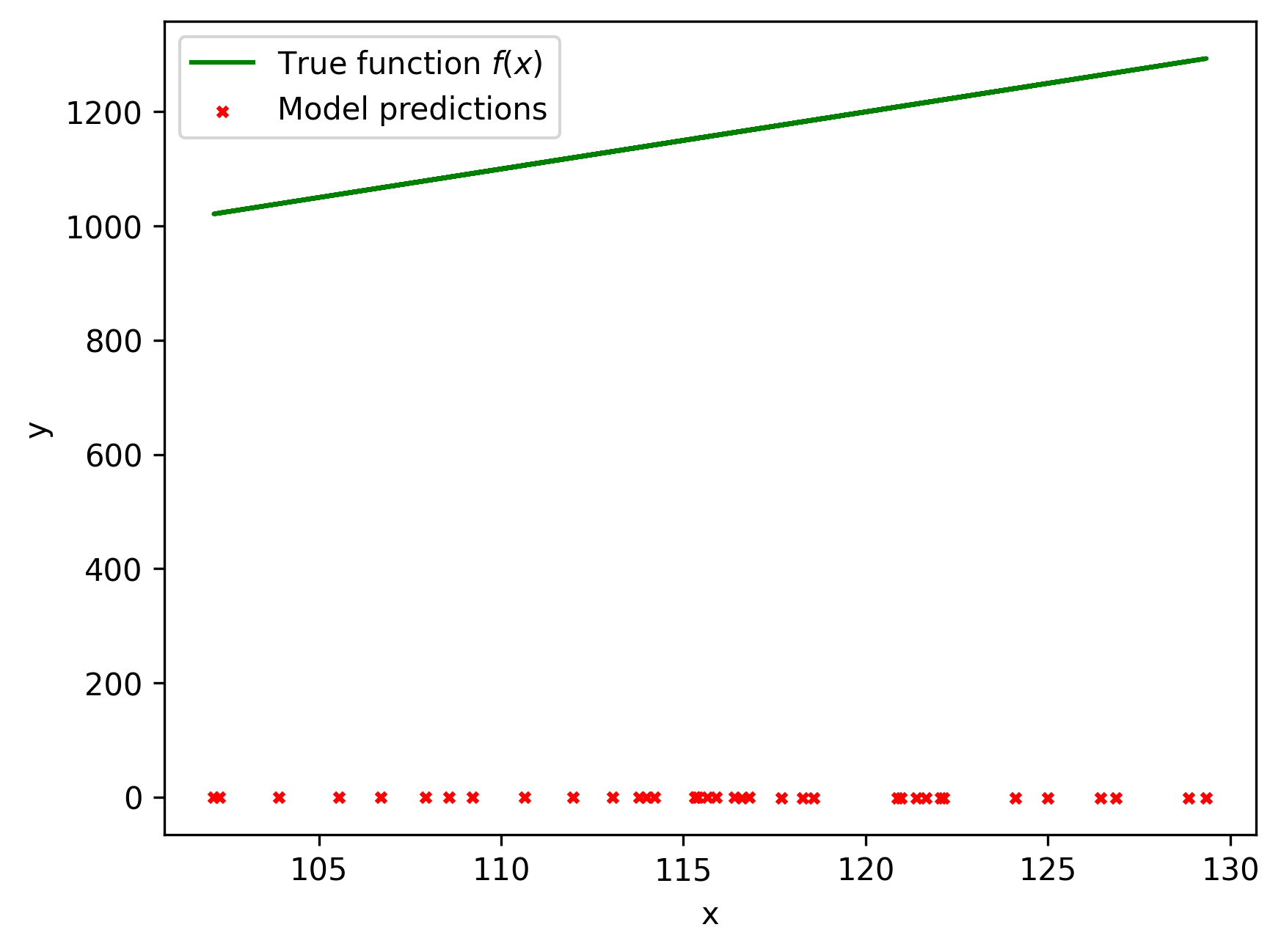}
\caption{Boundary values emerging as inputs move beyond the training range. Left: inputs within the training range; the model tracks the true function accurately. Center and right: as inputs grow, predictions saturate at fixed limits $(B^-, B^+)$ regardless of the true function value or the in-context examples provided.}
\label{sequence}
\end{figure}

\subsection{Final Normalization Bounds the Readout}
\label{sec:normalizationA}

The observed boundary values suggest that the model's predictions are constrained to a finite range. A natural hypothesis is that normalization causes this constraint. We show that this is partly correct: normalization induces a hard output bound when the decoder reads a final normalized representation. However, internal normalization layers alone do not imply bounded predictions, because residual connections can carry unnormalized information through the network.

\paragraph{Internal normalization does not bound the residual stream.} Consider a pre-norm Transformer with $L$ blocks, sequence length $n$, and hidden dimension $d$. Let $X_\ell \in \mathbb{R}^{n\times d}$ denote the residual stream before block $\ell+1$, for $\ell=0,\ldots,L-1$, with $X_0$ the embedded input sequence. Each block consists of an attention sublayer followed by an MLP sublayer, both applied in pre-norm form. Define the maps
\[
    \phi_{\ell+1}
    :=
    \mathrm{Attn}_{\ell+1}
    \circ
    \mathrm{LN}^{\mathrm{attn}}_{\ell+1},
    \qquad
    \psi_{\ell+1}
    :=
    \mathrm{MLP}_{\ell+1}
    \circ
    \mathrm{LN}^{\mathrm{mlp}}_{\ell+1},
\]
where
$\mathrm{LN}^{\mathrm{attn}}_{\ell+1},
\mathrm{LN}^{\mathrm{mlp}}_{\ell+1}:
\mathbb{R}^{n\times d}\to\mathbb{R}^{n\times d}$
act tokenwise on the residual stream, and
$\mathrm{Attn}_{\ell+1},\mathrm{MLP}_{\ell+1}:
\mathbb{R}^{n\times d}\to\mathbb{R}^{n\times d}$ include the usual output projections. The pre-norm block is
\begin{align}
    \widetilde X_\ell
        &=
        X_\ell + \phi_{\ell+1}(X_\ell),
        \label{eq:prenorm-attn}\\
    X_{\ell+1}
        &=
        \widetilde X_\ell
        +
        \psi_{\ell+1}(\widetilde X_\ell).
        \label{eq:prenorm-mlp}
\end{align}
Substituting \eqref{eq:prenorm-attn} into
\eqref{eq:prenorm-mlp} gives
\begin{equation}
    X_{\ell+1}
    =
    X_\ell
    +
    \phi_{\ell+1}(X_\ell)
    +
    \psi_{\ell+1}\!\left(
        X_\ell+\phi_{\ell+1}(X_\ell)
    \right).
    \label{eq:prenorm-expanded}
\end{equation}
Equivalently, defining the block increment
\[
    \Phi_{\ell+1}(X)
    :=
    \phi_{\ell+1}(X)
    +
    \psi_{\ell+1}\!\left(
        X+\phi_{\ell+1}(X)
    \right),
\]
we obtain the residual recursion
\begin{equation}
    X_{\ell+1}
    =
    X_\ell+\Phi_{\ell+1}(X_\ell),
    \qquad
    \ell=0,\ldots,L-1.
    \label{eq:prenorm-residual}
\end{equation}
Unrolling over the $L$ blocks yields
\begin{equation}
    X_L
    =
    X_0
    +
    \sum_{\ell=1}^{L}
    \Phi_\ell(X_{\ell-1}).
    \label{eq:prenorm-unrolled}
\end{equation}

Equation~\ref{eq:prenorm-unrolled} shows that internal normalization does not, by itself, bound the residual stream. In particular, any unbounded component present in the embedded input $X_0$ can be carried through the network by the residual connection. Although the attention and MLP sublayers are applied to normalized representations, their outputs are added to the residual stream rather than replacing it. Thus, the representation $X_L$ read by the decoder is not uniformly bounded merely because the sublayers are internally normalized.

\paragraph{Final normalization creates a bounded readout.}
The situation changes when the decoder reads a final normalized representation. In standard transformer architectures, the prediction at the query position is produced after a final normalization:
\[
    \hat g(x)
    =
    \mathrm{Dec}\!\left(
        \mathrm{Norm}_{\mathrm{final}}(X_L(x))_q
    \right),
\]

where $q$ denotes the query token  and Dec is the decoder. Unlike internal normalization, this normalization applies directly to the representation read by the decoder. Therefore, if $\mathrm{Norm}_{\mathrm{final}}$ maps every query representation into a fixed compact set and $\mathrm{Dec}$ is continuous, the prediction is bounded.

The following lemma makes this compactness statement explicit.

\begin{lemma}[Boundedness induced by LayerNorm]
\label{lemma:compact}
For any hidden representation $z\in\mathbb{R}^d$, LayerNorm with
parameters $\gamma,\beta\in\mathbb{R}^d$,
\[
\mathrm{LN}(z)
=
\gamma\odot
\frac{z-\mu(z)\mathbf{1}}
{\sqrt{\sigma^2(z)+\varepsilon}}
+
\beta,
\]
satisfies $\mathrm{LN}(z)\in\mathcal{K}_d$ for all
$z\in\mathbb{R}^d$, where
\[
\mathcal{K}_d
=
\left\{
y\in\mathbb{R}^d:
\|y\|_2
\le
\|\gamma\|_\infty\sqrt{d}+\|\beta\|_2
\right\}.
\]
In particular, $\mathcal{K}_d$ is compact and depends only on $d$ and the LayerNorm parameters. For RMSNorm, the same conclusion holds with the corresponding scale-normalized compact set.
\end{lemma}

\begin{theorem}[Boundedness from final normalization]
\label{thm:boundedness}
Let $\hat g(x)$ be the scalar prediction of a Transformer whose decoder is applied to a final normalized query representation:
\[
    \hat g(x)
    =
    \mathrm{Dec}\!\left(
        \mathrm{Norm}_{\mathrm{final}}(X_L(x))_q
    \right),
\]
where $\mathrm{Dec}:\mathbb{R}^d\to\mathbb{R}$ is continuous. Then there exists a constant $B>0$ such that
\[
    |\hat g(x)|\le B
    \qquad
    \text{for all inputs } x .
\]
\end{theorem}

\noindent
Proofs are given in Appendix~\ref{app:proofTh1}. The theorem is a statement about the representation read by the decoder: final normalization bounds the decoder input, whereas internal normalization does not by itself bound the residual stream. As a direct consequence, over an unbounded domain, such a model cannot satisfy $\mathrm{ICL}_2$ for any non-constant polynomial target. Indeed, the model's predictions are uniformly bounded, while any non-constant polynomial has unbounded range; see Appendix~\ref{app:proofNOICL2}.

Theorem~\ref{thm:boundedness} explains the bounded output range associated with boundary values. It does not, however, establish convergence to particular limits; in Section~\ref{sec:attention-bottleneck}, we identify a complementary mechanism that helps explain this convergence.

\subsection{Removing the Bound Does Not Restore ICL$_2$}\label{sec:no-equivariance}

%\paragraph{Removing final normalization.}

If final normalization were the only obstacle, the model might already compute the correct regression rule internally, with normalization merely constraining its output. To test this hypothesis and look at this internal solution, we removed final normalization from the trained baseline model only at test time, leaving all learned parameters unchanged. The resulting predictions are no longer subject to the final-normalization bound and can reach large magnitudes, but the error increases dramatically even in-distribution (Table~\ref{tab:final_norm}).

We interpret this post-hoc intervention cautiously, because the decoder was trained to operate on normalized representations. We performed a cleaner architectural test by training an otherwise identical model from scratch without final normalization. The resulting model achieves excellent in-distribution performance and is no longer constrained by a normalized readout, yet its error still grows sharply under coefficient scaling (Table~\ref{tab:final_norm}). Thus, final normalization explains the bounded readout, but it is not the sole cause of the failure of $\mathrm{ICL}_2$.  Further controls that enlarge the final-normalization scale lead to the same conclusion (Appendix ~\ref{app:final_norm_scale})

\begin{table}[!ht]
\centering
\small
\resizebox{\textwidth}{!}{
\begin{tabular}{lcccccccccc}
\toprule
& $\sigma=1$ & $2$ & $3$ & $4$ & $5$ & $6$ & $7$ & $8$ & $9$ & $10$ \\
\midrule
Final norm
& $9{\times}10^{-6}$ & $0.04$ & $0.62$ & $1.95$ & $4.29$
& $7.14$ & $10.36$ & $15.79$ & $24.48$ & $27.06$ \\

Norm removed at test time
& $434$ & $1239$ & $1676$ & $1646$ & $1734$
& $1730$ & $1716$ & $1656$ & $1609$ & $1558$ \\

No final norm (retrained)
& $4.4{\times}10^{-6}$ & $0.025$ & $0.63$ & $1.65$ & $5.11$
& $6.67$ & $12.18$ & $15.82$ & $25.15$ & $35.98$ \\
\bottomrule
\end{tabular}
}
\caption{
Effect of final normalization on degree-1 polynomial regression under distribution shift. \emph{Norm removed at test time} denotes the standard trained model with final normalization replaced by the identity only during inference; \emph{no final norm} denotes an otherwise identical model trained from scratch without final normalization.
}
\label{tab:final_norm}
\end{table}

\paragraph{Label-scale equivariance.}

We next ask whether the learned predictor is equivariant to rescaling the in-context labels. We test both models trained from scratch, with and without final normalization. Given a prompt $(x_1,g(x_1),\ldots,x_p,g(x_p),x)$, we multiply all context labels and the target by a factor $\alpha$. For polynomial regression, this is equivalent to scaling all coefficients by $\alpha$. A scale-equivariant regression rule, such as least squares, should therefore scale its prediction by the same factor.

We measure the MSE against the scaled target $\alpha g(x)$ and the scale-response ratio

\[
r(\alpha)
=
\frac{
\mathbb{E}\bigl[
|\hat f_\theta(x \mid \{x_i,\alpha g(x_i)\})|
\bigr]
}{
\alpha\,
\mathbb{E}\bigl[
|\hat f_\theta(x \mid \{x_i,g(x_i)\})|
\bigr]
}.
\]
For a scale-equivariant predictor, $r(\alpha)\approx 1$ for all $\alpha$; values below one indicate that predictions grow more slowly than the label scale.

\begin{figure}[!ht]
    \centering
    \includegraphics[width=\linewidth]{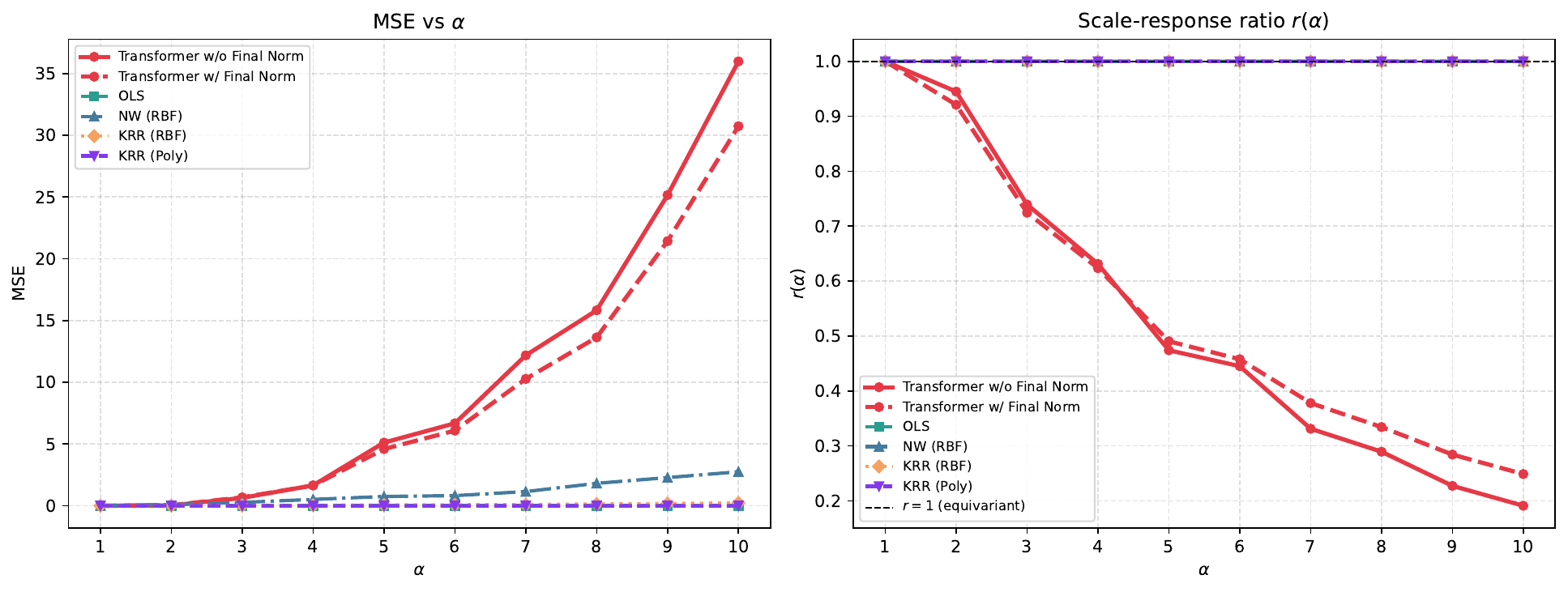}
    \caption{Label-scale equivariance test on degree-1 regression.
    Inputs and coefficients are sampled from the training distribution $\mathcal{U}(-1,1)$, but at evaluation time all in-context labels and target labels are multiplied by $\alpha$. Left: MSE against the scaled target $\alpha g(x)$. Right: scale-response ratio $r(\alpha)$, where $r(\alpha)=1$ corresponds to exact scale equivariance.}
    \label{fig:label_scale}
\end{figure}

Figure~\ref{fig:label_scale} shows that transformers do not exhibit the scale equivariance expected of an in-context regression algorithm. The final-normalized model degrades as $\alpha$ increases, consistent with readout-side saturation. More importantly, the model without final normalization behaves similarly: its MSE grows sharply and $r(\alpha)$ falls well below one as $\alpha$ increases. In contrast, least squares and kernel ridge regression remain close to $r(\alpha)=1$. Thus, removing the output bound does not reveal an underlying scale-equivariant regression algorithm; the model also fails to use the in-context labels in a scale-consistent way.

Together, these results separate boundary-value saturation from the broader failure of $\mathrm{ICL}_2$. Final normalization creates a hard readout-side bottleneck, but removing or rescaling that bottleneck does not restore distribution-independent in-context learning. This motivates examining other architectural components.

\section{Attention as a Context Bottleneck}
\label{sec:attention-bottleneck}

The previous section shows that removing or rescaling the final normalization bottleneck does not restore $\mathrm{ICL}_2$. Does the failure persist when other standard Transformer components are removed?

To answer this question, we perform a sequence of architectural ablations, progressively simplifying the standard Transformer block. The failure persists after removing MLP blocks, and even in attention-only models. These models still achieve strong in-distribution ICL, yet degrade sharply under distribution shift (Appendix~\ref{app:attn_only_results}). Thus, attention alone is sufficient for this failure to persist. We next examine how attention changes under distribution shift.

\paragraph{How attention can become a context bottleneck.}

For ICL, the model must aggregate information from the examples in the prompt to predict the output for a new query. In a Transformer, this interaction is handled by self-attention. Consider one attention head at the query position. Let $q\in\mathbb{R}^{d_h}$ denote the query vector and $k_j,v_j\in\mathbb{R}^{d_h}$ the key and value vectors associated with context position $j\in\{1,\ldots,n\}$. The attention output is
\[
\operatorname{Attn}(q)
=
\sum_{j=1}^{n} a_j v_j,
\qquad
a_j
=
\frac{\exp(z_j)}
{\sum_{r=1}^{n}\exp(z_r)},
\qquad
z_j
=
\frac{q^\top k_j}{\sqrt{d_h}},
\]
where $z_j$ is the query--key score and $a_j$ the corresponding attention weight.

Let $j^\star=\arg\max_j z_j$, and suppose that the largest score is separated from all others by at least $\Delta\geq 0$:
\[
z_{j^\star}-z_j\geq\Delta
\qquad
\text{for every } j\neq j^\star.
\]
Such a $\Delta$ can be taken as the gap between the largest and second-largest query--key scores. Then
\[
a_{j^\star}
\geq
\frac{1}{1+(n-1)e^{-\Delta}},
\qquad
a_j
\leq
e^{-\Delta}a_{j^\star}
\quad
\text{for every } j\neq j^\star.
\]

Thus, as the score gap grows, attention assigned to the remaining context positions is exponentially suppressed relative to the dominant one. Attention can therefore become concentrated on only a small part of the prompt, potentially limiting the context-dependent adaptation required for ICL.

The attention-only ablation shows that the failure persists in an attention-only architecture, but does not establish whether this concentration mechanism is active under the distribution shifts where performance degrades. We now test this directly.

\begin{figure}[!ht]
    \centering
    \includegraphics[width=\linewidth]{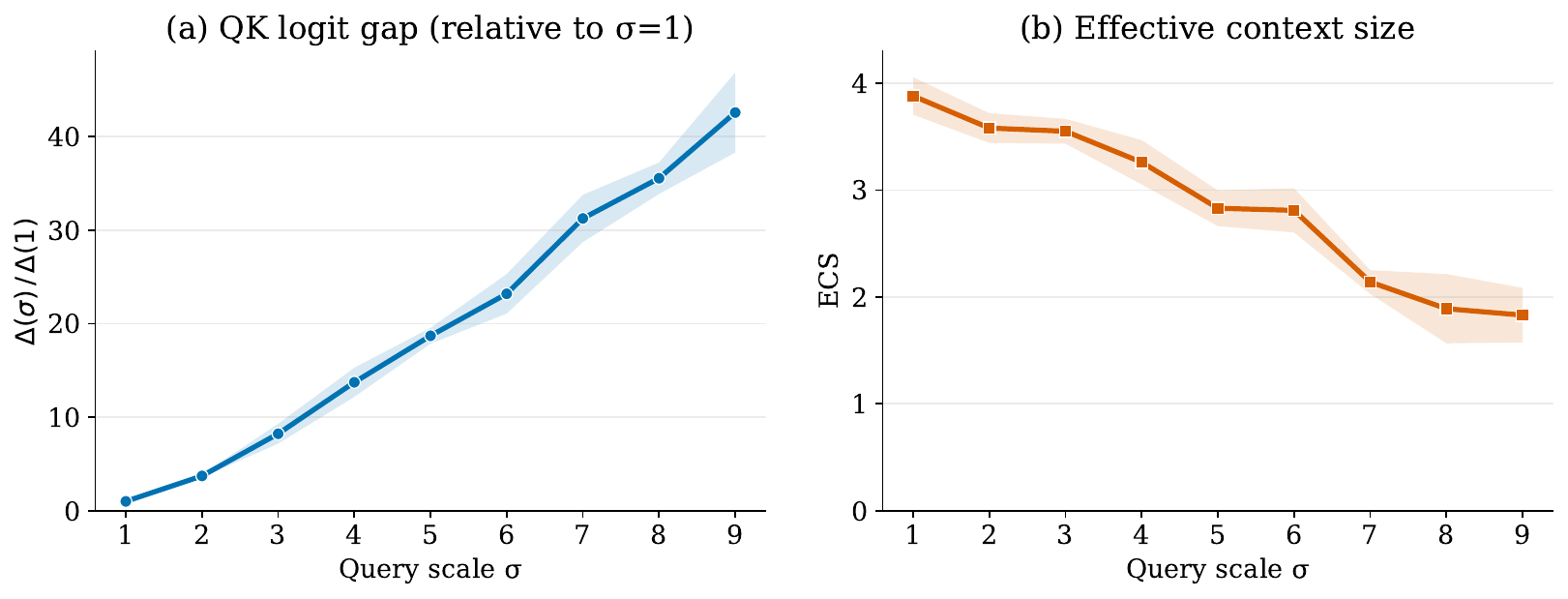}
    \caption{ \textbf{First-layer attention diagnostics under distribution shift} (query $\sim \mathcal{U}(-\sigma,\sigma)$, context in-distribution; degree-1 model, mean $\pm$ std over 5 seeds). \textbf{(a)} Relative first-layer query--key score gap. The gap $\Delta(\sigma)$ between the largest and second-largest scores at the query position is normalized by its value at $\sigma=1$; it grows as the query leaves the training support. \textbf{(b)} Effective context size (ECS) at query position, which decreases correspondingly as attention concentrates on fewer tokens.
    }
    \label{fig:qk_gap_ratio}
\end{figure}

\paragraph{The effective context shrinks under distribution shift.} 
We first measure the separation between query--key scores at the query position under distribution shift. For each attention head, let
\[
\Delta = z_{(1)}-z_{(2)}
\]
denote the gap between the largest and second-largest scores. This gap increases sharply as the input distribution shifts farther from its training support (Figure~\ref{fig:qk_gap_ratio}). As shown above, a larger $\Delta$ exponentially suppresses the attention assigned to non-dominant positions relative to the dominant one, resulting in increasingly concentrated attention weights.

We next quantify this concentration using the effective context size (ECS) \[ \mathrm{ECS}(a)=\frac{1}{\sum_{j=1}^{n} a_j^2},\] where $a=(a_1,\ldots,a_n)$ is the attention-weight vector over the $n$ positions available to the query. Uniform attention gives $\mathrm{ECS}=n$, whereas $\mathrm{ECS}=1$ when all attention mass is concentrated on a single token. Both $\Delta$ and ECS are computed separately for each attention head; we report averages over heads, evaluation samples, test functions, and random seeds. For the degree-1 model, the first-layer mean ECS decreases under distribution shift, reaching approximately $1.8$ at $\sigma=9$. Thus, although the full prompt remains available, the effective context used by the query shrinks as the distribution moves farther from training.

\paragraph{Causal intervention on attention.}
The preceding results suggest that attention becomes increasingly concentrated under distribution shift. We then ask whether reducing this concentration can mitigate the OOD failure. We test this in two complementary ways.
First, we intervene on a trained model at inference time by temperature-scaling the query--key scores, $z \mapsto z/T$. Here, $T=1$ recovers the original attention distribution, while larger $T$ progressively flattens it. Post-hoc smoothing does not rescue OOD performance, while stronger smoothing increases the error at large shifts (Table~\ref{tab:attention_smoothing}).

Because this intervention changes the attention maps after training, the model has no opportunity to adapt to the flattened attention distribution. We therefore also train models from scratch with temperature-scaled softmax attention. Although these models adapt throughout training to the modified attention rule, they still exhibit the same qualitative degradation under distribution shift. Thus, simply reducing the sharpness of attention, either post hoc or throughout training, is insufficient to recover $\mathrm{ICL}_2$ (Table~\ref{table:LF}).

We extend this retraining analysis to alternative attention mechanisms, including linear attention, sparsemax \citep{martins2016softmax}, entmax \citep{peters2019sparse}, and CosFormer-style attention \citep{qin:etal:2022}. Their quantitative OOD behavior differs, but the same qualitative failure under distribution shift persists across all variants (Table~\ref{table:LF}).

Finally, we complement these experiments with an asymptotic analysis in Appendix~\ref{sec:mathproofs}. In the large-query regime, the leading attention update is governed by the query and learned model parameters rather than by the in-context examples. Combined with the final normalized readout analyzed in Section~\ref{boundaryvalues}, this helps explain the empirical convergence toward model-specific boundary values, beyond the boundedness guaranteed by final normalization alone.

\section{Conclusion}

We have shown that strong in-distribution ICL does not imply distribution-independent ICL in controlled settings where classical estimators can generalize exactly. Transformers achieve near-zero in-distribution error, but degrade sharply under distribution shifts.

We identify two architectural bottlenecks. First, final normalization creates a bounded readout that explains boundary values; removing it eliminates hard output bounds but does not restore $\mathrm{ICL}_2$, so the model is not merely clipping a correct internal rule. Second, attention can become a context bottleneck: under distribution shift, the query update may depend on only a small part of the prompt, limiting the context-dependent adaptation required for ICL.

Our analysis is primarily focused on decoder-only transformers with relatively modest model sizes. Appendix \ref{app:architectural_interventions} reports complementary experiments with architectures that incorporate additional inductive biases.  Increasing model scale may affect quantitative aspects of the behavior. However, scale alone does not remove the readout-side limitation when predictions are decoded from a final normalized representation, nor does it directly address the attention concentration mechanism.

Overall, these results help clarify the mechanisms and limits of ICL in transformers. Strong $\mathrm{ICL}_1$ should not be taken as evidence of a distribution-independent learning procedure, and ICL-trained predictors should be used with caution as substitutes for retraining, especially on real data where test-time distributions may shift.

\begin{ack}
We are grateful for the support of the ANITI IA Cluster (ANR-23-IACL-0002), the SARER and Summ-RE projects (ANR-20-CE23-0017), and the COCOBOTS project (ANR-21-FAI2-0005). We also thank CALMIP for providing computational resources under grant 2016-P23060.
\end{ack}

%%%%%%%%%%%%%%%%%%%%%%%%%%%%%%%%%%%%%%%%%%%%%%%%%%%%%%%%%%%%
\bibliography{custom}
\bibliographystyle{icml2026}

\appendix

\newpage
\section{Limitations}

Our experiments use a controlled setting with synthetic mathematical functions. Although this setup enables a precise analysis of model behavior and architectural effects, it abstracts away from many complexities present in real-world NLP tasks. We use polynomial functions as our main testbed to identify architectural mechanisms that can limit ICL generalization in transformers under distribution shift. 

Despite these limitations, our framework and findings lay the groundwork for understanding and addressing ICL failure modes, offering a foundation for improved architectural design and training strategies in both synthetic and natural language domains.  In addition, small Transformers are increasingly being used as in-context predictors, making it important to understand their behavior and limitations under distribution shift.

\section{Related Work}

\paragraph{In-context learning and the algorithmic inference hypothesis.}
Small transformers trained from scratch on synthetic function classes have become the standard paradigm for studying ICL mechanisms~\citep{garg:etal:2022,akyurek:etal:2022,von-oswald:etal:2022}. Since the introduction of ICL by \citet{brown:etal:2020}, a central hypothesis has been that transformers implement learning algorithms at inference time. Several works support this view by showing that transformers \emph{can} represent optimization procedures such as gradient descent or least-squares regression under suitable constructions~\citep{von-oswald:etal:2022,akyurek:etal:2022, mahankali2023one,dai2023can}. Complementary empirical results show that transformers trained from scratch on regression tasks match classical estimators under matched distributions~\citep{garg:etal:2022,li2023transformers}, and recent work relates ICL to kernel methods such as Nadaraya-Watson regression~\citep{han2025understanding,shen2025understanding}.%, showing that attention can approximate such estimators in distribution. These findings motivate the algorithmic inference hypothesis, but they are established exclusively under matched distributions. We show that the hypothesis fails as soon as distributions shift.

\paragraph{Distribution shift and limitations of ICL.}
A growing body of work shows that ICL performance can degrade when the inference-time distribution differs from training~\citep{zhang:etal:2024,giannou:etal:2024, naim:asher:2024b}. \citet{naim:asher:2024b} systematically characterize distribution shifts causing significant degradation for linear functions and introduce the notion of boundary values; we extend their analysis to polynomial classes of degrees 1--6 and show that boundary values arise when predictions are decoded from a final normalized representation. Notably, the theoretical analyses of \citet{zhang:etal:2024} and \citet{giannou:etal:2024} rely on linear attention, whereas we study standard softmax transformers as commonly deployed. Existing work largely documents performance degradation without identifying whether failures arise from training procedures or architectural constraints. We address this by testing a range of training interventions and identifying two architectural mechanisms that help explain the observed failure.

%\paragraph{Mechanistic interpretations of transformers.} Several lines of work attempt to explain ICL through mechanistic perspectives. Some studies attribute transformer behavior to memorization and pattern matching via attention~\citep{geva:etal:2021,geva:etal:2023,yu:etal:2023, bietti:etal:2024}, while others interpret it through Bayesian frameworks~\citep{xie2021explanation}. These perspectives are primarily validated under matched distributions and do not speak to the architectural origins of failure under distribution shift, which is the focus of our theoretical results.

\paragraph{Task-specific transformers in real-world systems.}
Small transformers trained from scratch for in-context prediction  initially emerged as diagnostic tools for studying ICL in controlled settings~\citep{garg:etal:2022, raventos2024pretraining, bhattamishra2023understanding, naim2025scaled}. They have increasingly become practical predictors in their own right, notably in tabular learning~\citep{tabpfn2022,contexttab2025}, reinforcement learning~\citep{wu2025mixture}, and scientific estimation~\citep{seifner2025context}. In these real-world deployments, distribution shift is not an edge case but a routine occurrence: the conditions at inference time frequently differ from those seen during training, whether due to domain drift, changing environments, or new data regimes. This shift from diagnostic tool to deployed system makes understanding ICL failure modes practically urgent: errors under distribution shift may be systematic, silent, and difficult to detect without the kind of architectural diagnosis we provide here.

\section{Training details}
\label{sec:trainingdetails}

 This section describes the default setup used for the polynomial-regression experiments; modifications for the higher-dimensional nonlinear setting and representation ablations are described in their respective sections.

\paragraph{Architecture.}
All experiments use a decoder-only Transformer with the same basic configuration as GPT-style models: 12 layers, 8 attention heads, and hidden dimension 256. The model is trained from scratch; no pretrained weights are used. Dropout is disabled, since every optimization step uses freshly sampled prompts. Each scalar input or output token is first embedded into the model dimension, the resulting sequence is processed autoregressively by the Transformer, and the hidden states are decoded back to scalar predictions. Our implementation builds on the code released by \citet{garg:etal:2022}.

\paragraph{Input and output representation.}
Scalar inputs and labels are represented using learned linear maps. In particular, each scalar token is mapped to $\mathbb{R}^{256}$ by a learned encoder $W_{\mathrm{enc}}$, and the Transformer outputs hidden vectors in the same space. Scalar predictions are obtained by applying a separate learned linear readout $W_{\mathrm{dec}}$ to the corresponding hidden state. 

\paragraph{Training Procedure.}
We train for 500k steps with Adam~\citep{kingma2014adam} (learning rate $10^{-4}$, batch size 64). Prompts are generated on the fly, so every step sees fresh data: we draw $g \sim \mathcal{D}_\mathcal{F}$, sample $x_1, \dots, x_{k+1}$ i.i.d. from $\mathcal{D}_\mathcal{I}$, and build the prompt from the evaluations of $g$ at these points.

\paragraph{Curriculum Learning.}
Models trained from scratch use a curriculum over the number of in-context examples. Training starts with prompts containing 11 input--output pairs, and the prompt length is increased by 2 every 2,000 optimization steps until reaching the maximum length $k=41$.
For models trained on multiple polynomial degrees, we additionally use a degree curriculum. In these runs, the model is trained sequentially on polynomial classes in increasing degree order, so that lower-degree tasks are introduced before higher-degree tasks. For each multi-degree experiment, we report the specific degree sequence used.

\paragraph{ICL Evaluation.}
We evaluate in-context learning (ICL) performance on held-out test distributions. For each testing scenario $(D_{\mathcal I}^{\mathrm{test}}, D_{\mathcal F}^{\mathrm{test}})$ and each random seed $s \in \mathcal S$, where $\mathcal S=\{42,123,456,789,1011\}$, we sample $N=100$ functions $g \sim D_{\mathcal F}^{\mathrm{test}}$. For each function, we sample $N_b=64$ batches, each containing $N_p=41$ input points drawn i.i.d. from $D_{\mathcal I}^{\mathrm{test}}$.

Within each batch $b$, the model is prompted with sequences of the form
\[
(x_1^b, g(x_1^b), \dots, x_{k-1}^b, g(x_{k-1}^b), x_k^b),
\quad k \ge 2,
\]
and predicts the corresponding output for $x_k^b$. For polynomial target functions $g \in \mathbb{R}_n[X]$, the first $n+1$ prediction positions in each batch are excluded from evaluation, since at least $n+1$ input--output pairs are required to uniquely determine the function.

\begin{equation}
\epsilon_\sigma^{(s)} =
\frac{1}{N}
\sum_{i=1}^{N}
\frac{1}{N_b}
\sum_{b=1}^{N_b}
\left(
\frac{1}{N_p - (n+1)}
\sum_{k=n+2}^{N_p}
\bigl(\mathrm{pred}_{i,k}^b - y_{i,k}^b\bigr)^2
\right),
\end{equation}

\noindent
where $n$ denotes the degree of the polynomial. The final reported metric is obtained by averaging over seeds:
\begin{equation}
\epsilon_\sigma =
\frac{1}{|\mathcal S|}
\sum_{s \in \mathcal S}
\epsilon_\sigma^{(s)}.
\end{equation}

To ensure fair and meaningful comparisons across models, we fix the random seed when generating the 100 test functions and the corresponding prompting points $x_i$. This guarantees that all models are evaluated on the same set of functions and input distributions. The purpose of this evaluation setup is to assess how well models generalize when progressively exposed to OOD data. By keeping the test conditions constant while varying the models, we can isolate the effect of model architecture and training on their ability to adapt to novel inputs through in-context learning.

\paragraph{Kernel baseline hyperparameters.}
All kernel baseline hyperparameters were fixed before evaluation. For the RBF kernel, we used
\[
k_{\mathrm{RBF}}(x,x')
=
\exp\!\left(-\frac{\|x-x'\|^2}{2\ell^2}\right),
\qquad
\ell = c_\ell\,\mathrm{std}(x)+\epsilon,
\]
with $c_\ell=1.0$ and $\epsilon=10^{-6}$. Here, $\mathrm{std}(x)$ is computed from the fixed input grid sampled under the training distribution. For KRR, the ridge parameter was set to $\lambda=10^{-3}$ for both the RBF and polynomial kernels. No hyperparameter tuning was performed on the out-of-distribution test conditions.

\paragraph{Computational Resources.} 
All models were trained on a single Nvidia A-100 80GB GPU, with an average training time of 4 hours.

\section{Evaluating Models Robustness and Generalization}\label{tables}

\begin{table}[!ht]
\centering
\resizebox{1\linewidth}{!}{
\begin{tabular}{|c|c|c|c|c|c|c|c|c|c|c|}
\hline
\textbf{Deg} & \boldmath{$\sigma=1$} & \boldmath{$\sigma=2$} & \boldmath{$\sigma=3$} & \boldmath{$\sigma=4$} & \boldmath{$\sigma=5$} & \boldmath{$\sigma=6$} & \boldmath{$\sigma=7$} & \boldmath{$\sigma=8$} & \boldmath{$\sigma=9$} & \boldmath{$\sigma=10$} \\
\hline
1 & $(9.0 \pm 1.0) \times 10^{-6}$ & $0.04 \pm 0.00$ & $0.6 \pm 0.1$ & $2.0 \pm 0.2$ & $4.3 \pm 0.2$ & $7.1 \pm 0.5$ & $10.4 \pm 0.8$ & $15.8 \pm 1.1$ & $24.5 \pm 2.1$ & $27.1 \pm 2.3$ \\
\hline
2 & $(5.1 \pm 3.2) \times 10^{-5}$ & $0.04 \pm 0.01$ & $0.5 \pm 0.1$ & $2.1 \pm 0.2$ & $5.3 \pm 0.6$ & $7.7 \pm 0.7$ & $12.6 \pm 0.8$ & $18.3 \pm 1.2$ & $24.8 \pm 1.1$ & $32.0 \pm 2.7$ \\
\hline
3 & $(1.8 \pm 0.9) \times 10^{-4}$ & $0.02 \pm 0.00$ & $0.4 \pm 0.1$ & $1.5 \pm 0.2$ & $3.7 \pm 0.3$ & $6.3 \pm 0.7$ & $10.8 \pm 1.0$ & $16.3 \pm 1.3$ & $22.5 \pm 2.7$ & $29.0 \pm 1.3$ \\
\hline
4 & $(3.8 \pm 1.0) \times 10^{-4}$ & $0.04 \pm 0.00$ & $0.4 \pm 0.1$ & $1.4 \pm 0.2$ & $3.7 \pm 0.2$ & $6.6 \pm 0.6$ & $11.4 \pm 1.1$ & $16.0 \pm 1.4$ & $24.1 \pm 2.3$ & $31.5 \pm 1.5$ \\
\hline
5 & $(6.4 \pm 2.7) \times 10^{-4}$ & $0.02 \pm 0.00$ & $0.3 \pm 0.1$ & $1.5 \pm 0.2$ & $3.5 \pm 0.1$ & $7.4 \pm 0.5$ & $13.2 \pm 0.9$ & $17.7 \pm 1.3$ & $25.4 \pm 3.7$ & $32.9 \pm 3.2$ \\
\hline
6 & $(9.1 \pm 3.7) \times 10^{-4}$ & $0.03 \pm 0.01$ & $0.4 \pm 0.1$ & $2.1 \pm 0.3$ & $4.4 \pm 0.8$ & $9.2 \pm 0.8$ & $14.7 \pm 1.9$ & $21.6 \pm 1.8$ & $29.1 \pm 2.6$ & $37.7 \pm 2.9$ \\
\hline
\end{tabular}
}
\caption{Evolution of the MSE for Transformer models with 12 layers and 8 attention heads, trained on polynomials of degree $i \in \{1,\dots,6\}$. During training, both input samples and coefficients are drawn from $\mathcal{U}(-1,1)$. At test time, input samples remain drawn from $\mathcal{U}(-1,1)$, while coefficients are drawn from $\mathcal{U}(-\sigma,\sigma)$ with $\sigma \in \{1,\cdots,10\}$. Each model is evaluated on its corresponding degree. Results are averaged over five random seeds (42, 123, 456, 789, 1011), with each seed evaluated on 100 functions, following the evaluation protocol described in Appendix \ref{sec:trainingdetails}.}
\label{table:3a}
\end{table}

\begin{figure}[!ht]
\center 
\includegraphics[width=\textwidth]{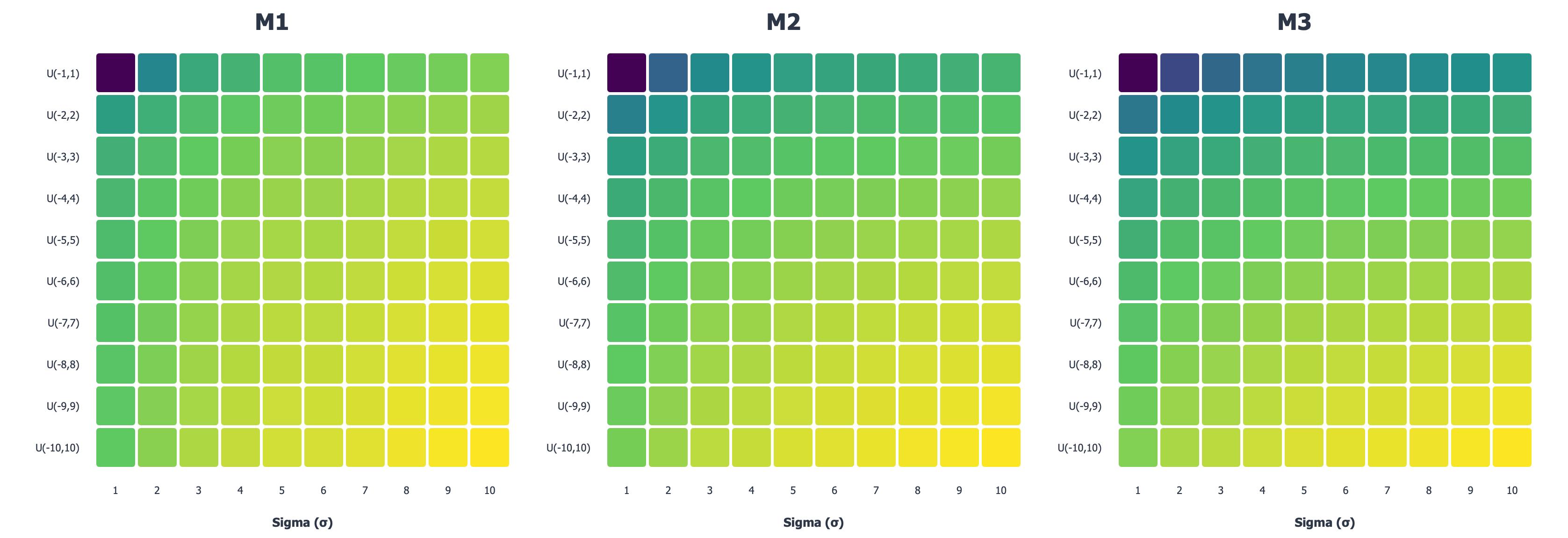}
\caption{Heatmaps of log MSE under joint input and coefficient shifts for models $M_1$, $M_2$, and $M_3$, trained on polynomial degrees 1, 2, and 3, respectively. Training inputs and coefficients are sampled from $\mathcal{U}(-1,1)$. At test time, inputs are sampled from $\mathcal{U}(-\sigma_I,\sigma_I)$ and coefficients from $\mathcal{U}(-\sigma_F,\sigma_F)$. The horizontal axis varies $\sigma_I$, and the vertical axis varies $\sigma_F$. Blue indicates lower error and yellow higher error.}
\label{hmap}
\end{figure}

\begin{table}[!ht]
\centering
\resizebox{1\linewidth}{!}{
\begin{tabular}{|c|c|c|c|c|c|c|c|c|c|c|}
\hline
\textbf{Deg} 
& \boldmath{$\sigma=1$} 
& \boldmath{$\sigma=2$} 
& \boldmath{$\sigma=3$} 
& \boldmath{$\sigma=4$} 
& \boldmath{$\sigma=5$} 
& \boldmath{$\sigma=6$} 
& \boldmath{$\sigma=7$} 
& \boldmath{$\sigma=8$} 
& \boldmath{$\sigma=9$} 
& \boldmath{$\sigma=10$} \\
\hline
1 & 2.05 $\times 10^{-5}$ & 0.02 & 0.15 & 0.27 & 0.37 & 0.44 & 0.52 & 0.56 & 0.61 & 0.65 \\
\hline
2 & 7.62 $\times 10^{-5}$ & 0.02 & 0.13 & 0.27 & 0.37 & 0.46 & 0.54 & 0.59 & 0.64 & 0.68 \\
\hline
3 & 2.53$\times 10^{-4}$ & 0.01 & 0.07 & 0.16 & 0.24 & 0.31 & 0.39 & 0.43 & 0.49 & 0.53 \\
\hline
4 & 1.35$\times 10^{-3}$ & 0.01 & 0.08 & 0.16 & 0.24 & 0.33 & 0.39 & 0.43 & 0.48 & 0.53 \\
\hline
5 & 2.32 $\times 10^{-3}$ & 0.01 & 0.06 & 0.15 & 0.24 & 0.32 & 0.39 & 0.43 & 0.50 & 0.56 \\
\hline
6 & 2.22 $\times 10^{-3}$ & 0.01 & 0.07 & 0.17 & 0.28 & 0.37 & 0.43 & 0.49 & 0.55 & 0.60 \\
\hline
\end{tabular}
}
\caption{NMSE for the models and coefficient-scaling experiments reported in Table~\ref{table:3a}. Each model is trained and evaluated on its corresponding polynomial degree. Entries are means over five evaluation seeds, following the protocol in Appendix~\ref{sec:trainingdetails}.}

\label{tab:nmse_generalization}
\end{table}

%\newpage

\section{Higher-Dimensional Nonlinear ICL}
\label{app:tabular}

\subsection{Experimental setup}
\label{app:tabular_setup}

To evaluate whether the observed failure extends beyond scalar polynomial regression, we consider a higher-dimensional nonlinear regression setting inspired by the Bayesian neural network component of TabPFN's synthetic prior. We use the same 12-layer, 8-head Transformer with hidden dimension 256 as in the main experiments.

For each training prompt, we independently sample a bias-free network with one ReLU hidden layer of width $h=100$. Within each prompt, we draw the inputs independently from $\mathcal{U}(-1,1)^{10}$ and use the same sampled network to generate all targets, without observation noise. We train the Transformer from scratch for approximately $500{,}000$ optimization steps with up to 100 in-context examples, using the same optimizer and learning rate as in Appendix~\ref{sec:trainingdetails}.

We evaluate two forms of distribution shift. In the \emph{target-scale shift}, the target output is multiplied by a scale factor $\sigma$ while the feature distribution remains fixed at $\mathcal{U}(-1,1)^{10}$. This provides an analogue of coefficient scaling in the polynomial experiments while leaving the input distribution unchanged.
In the \emph{feature-support shift}, the target function is held fixed while the input distribution is expanded from $\mathcal{U}(-1,1)^{10}$ to $\mathcal{U}(-\sigma,\sigma)^{10}$.

For the distribution-shift evaluation, we use five independently generated test draws with seeds $\{42,123,456,789,1011\}$. For a given test draw, the same underlying target function and the same base feature sample are reused across values of $\sigma$; only the target scale or feature scale is changed. This paired evaluation isolates the effect of distribution shift from variation caused by resampling the underlying test task.

Because the intrinsic output scale of the random-network prior varies across tasks and shifts, we report normalized mean squared error (NMSE),
\[
    \mathrm{NMSE}
    =
    \frac{
        \mathbb{E}\!\left[(y-\hat{y})^2\right]
    }{
        \mathbb{E}\!\left[y^2\right]
    }.
\]
Under this normalization, the zero predictor $\hat{y}=0$ has $\mathrm{NMSE}=1$. Values approaching one therefore correspond to performance approaching the zero-prediction baseline. 

\subsection{Distribution-shift degradation}
\label{app:tabular_shift}

Figure~\ref{fig:tabular_shift} shows the degradation under the two forms of distribution shift. In distribution ($\sigma=1$), the model achieves an NMSE of $0.229\pm0.020$. Under target-scale shift, performance degrades progressively as the target scale increases, reaching $0.792\pm0.022$ at $\sigma=10$. Feature-support shift produces a substantially earlier and sharper degradation: NMSE reaches $0.768\pm0.025$ already at $\sigma=3$ and then remains close to the zero-prediction baseline.

These results show that the qualitative distribution-shift degradation is not restricted to scalar polynomial regression, but also appears with higher-dimensional inputs and nonlinear target functions.

\subsection{Boundary-value behavior}
\label{app:tabular_boundaries}

We next examine whether the boundary-value behavior observed in the polynomial experiments also appears in this higher-dimensional setting. We sample 20 target-function/context pairs, each with 40 in-context examples drawn from the training support, and evaluate the same model as the query moves progressively farther outside that support along each of the ten coordinate axes, in both directions.

\begin{figure}[!ht]
    \centering
    \includegraphics[width=\textwidth]{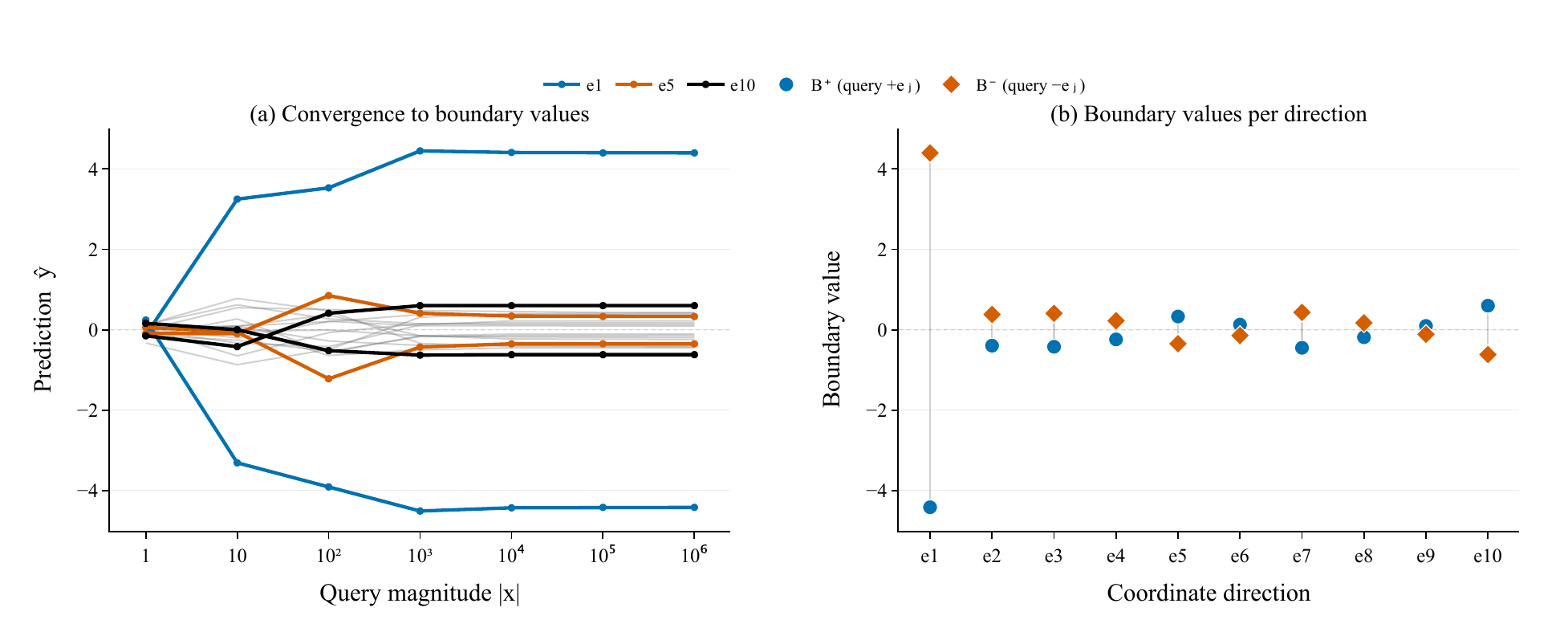}
\caption{Boundary-value behavior in the 10-dimensional nonlinear regression setting. Within each trial, 40 in-context examples remain fixed while the query moves along each coordinate axis in both directions. (a) Prediction trajectories as query magnitude increases, with axes $e_1$, $e_5$, and $e_{10}$ highlighted. (b) Estimated boundary values $B_j^{+}$ and $B_j^{-}$ for the positive and negative query directions. Results use 20 independently sampled target-function/context pairs and a fixed trained model.}
    \label{fig:tabular_boundaries}
\end{figure}

Figure~\ref{fig:tabular_boundaries} shows the resulting trajectories. Despite substantial variation near the training distribution, the predictions settle to finite values as the query moves progressively farther outside the training support. Different coordinate directions approach different asymptotes, and the limiting value also depends on the sign of the query direction.

\newpage 

\section{Two-Layer Attention-Only Models Enable ICL$_1$ Across Polynomial Degrees}
\label{app:attn_only_results}

\begin{table}[!ht]
\centering
\resizebox{\linewidth}{!}{
\begin{tabular}{|l|l|l|l|l|l|l|l|l|l|l|l|}
 \hline
% \rowcolor{orange!30}
\textbf{Deg} 
& \boldmath{$\sigma=1$} 
& \boldmath{$\sigma=2$} 
& \boldmath{$\sigma=3$} 
& \boldmath{$\sigma=4$} 
& \boldmath{$\sigma=5$} 
& \boldmath{$\sigma=6$} 
& \boldmath{$\sigma=7$} 
& \boldmath{$\sigma=8$} 
& \boldmath{$\sigma=9$} 
& \boldmath{$\sigma=10$} \\
 \hline
{1} & 0.00 & 0.03& 0.83& 2.12& 6.26& 7.60& 14.15& 18.07& 27.66& 39.17 \\
 \hline 

{2} &  0.00& 0.04& 1.07& 2.76& 7.23& 10.69& 15.70& 20.25& 28.85& 44.72 \\
 \hline 
{3} & 0.00& 0.05 & 0.86& 2.25& 6.37& 11.44& 15.05& 19.77& 29.68& 46.47 \\

 \hline %\\ [1ex] 
\end{tabular}
}

\caption{MSE under coefficient scaling for attention-only Transformers with 2 layers and 8 attention heads, trained separately on polynomial degrees 1--3. Training inputs and coefficients are sampled from $\mathcal{U}(-1,1)$. At test time, inputs remain sampled from $\mathcal{U}(-1,1)$, while coefficients are sampled from $\mathcal{U}(-\sigma,\sigma)$. Each model is evaluated on its training degree.}

\label{table:2L8AH}
\end{table}

\section{Training and Representation Interventions Do Not Remove the Failure}
\label{sec:rulingout}

This section provides the detailed results for the training and representation interventions summarized in Section~\ref{sec:failure}. We consider structured supervision, longer training, expanded training distributions, different degree schedules, input-support shifts, nonlinear encoder and decoder mappings, and prompt-wise label normalization. These interventions affect quantitative performance but do not remove the qualitative degradation under distribution shift.

\subsection{Comparison of Different Training Schemes}
\label{appendix:training-schemes}

A natural hypothesis is that providing more structure during training could encourage the model to learn a rule that generalizes beyond the training distribution.  For linear functions $f(x)=a+bx$, the function class is spanned by the canonical basis $(1,x)$. Rather than training only on a dense, unstructured distribution of coefficient pairs, we therefore test whether explicitly exposing the model to this structure improves generalization under distribution shift.

\begin{figure}[!ht]
    \centering
    \includegraphics[width=0.8\textwidth, height = 4.3cm]{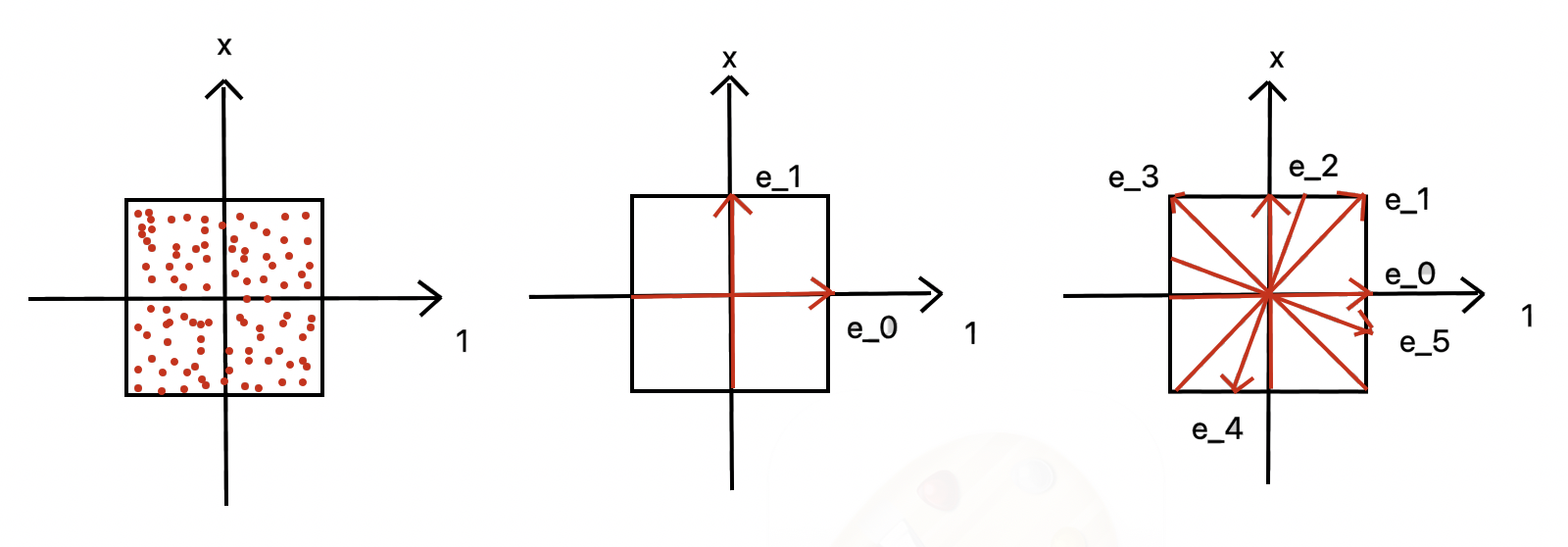}
    \caption{Illustration of training schemes for degree-1 polynomials. (Left) Standard training $\mathcal{T}$ on a dense cloud of points. (Middle) Basis-aligned training $\mathcal{T}_1$. (Right) Multi-directional structured training $\mathcal{T}_2$. The rectangle represents the set of polynomials of degree 1 with coefficients sampled from $\mathcal{U}(-1,1)$.}
    \label{polynomialfigures}
\end{figure}

We compare the standard training procedure $\mathcal{T}$ with two structured alternatives, $\mathcal{T}_1$ and $\mathcal{T}_2$ (Figure~\ref{polynomialfigures}).

In $\mathcal{T}_1$, the model is trained separately on the two
orthogonal components $a \cdot 1$ and $b \cdot x$, with $a$ and $b$ sampled from the same distribution. These components suffice to represent any linear function, but for neural networks, learning them separately does not guarantee effective composition for arbitrary linear combinations. $\mathcal{T}_1$ therefore tests whether structured basis-aligned supervision can produce a capable predictor for arbitrary linear functions.

In $\mathcal{T}_2$, the model is trained on multiple linear combinations of the basis vectors,
\[
  a \cdot e_i = a \cdot \bigl(\cos(\theta_i) \cdot 1
  + \sin(\theta_i) \cdot x\bigr),
\]
spanning a range of orientations in function space
(Figure~\ref{polynomialfigures}). This provides richer, structured supervision than $\mathcal{T}_1$ and tests whether exposure to multiple directions yields a stronger baseline predictor and improved robustness under coefficient-scale shift.

Table~\ref{table:training} summarizes performance under increasing distributional shifts in the coefficients. While $\mathcal{T}_2$ consistently outperforms $\mathcal{T}_1$, neither structured scheme matches the in-distribution accuracy of the standard unstructured procedure $\mathcal{T}$. In particular, both $\mathcal{T}_1$ and $\mathcal{T}_2$ already incur substantially larger error at $\sigma=1$,
before any coefficient-scale shift is applied. Therefore, their
performance under larger shifts should not be interpreted as an isolated failure of extrapolation.

Instead, this experiment shows that the structured, basis-aware training schemes considered here do not yield baseline predictors as capable as those obtained by standard unstructured training. Although exposing the model to multiple directions in function space improves over training on separate basis components alone, these interventions do not recover the
near-zero in-distribution performance of $\mathcal{T}$ and therefore do not resolve the observed generalization problem.

\begin{table}[!ht]
\centering
\resizebox{\linewidth}{!}{
\begin{tabular}{|l|l|l|l|l|l|l|l|l|l|l|}
\hline
models \ $\backslash$ \ $\sigma$ & 1 & 2 & 3 & 4 & 5 & 6 & 7 & 8 & 9 & 10 \\ 
 \hline
$\mathcal{T}$   & $9 \times 10^{-6}$& 0.04 & 0.62& 1.95& 4.29& 7.14& 10.36& 15.79& 24.48& 27.06 \\ 
 \hline
 $\mathcal{T}_1$  & 0.11& 1.16& 3.33& 5.08& 10.19& 12.56& 18.67& 23.38& 32.98& 44.91 \\ 
 \hline
 $\mathcal{T}_2$   & 0.01& 0.75& 2.71& 4.48& 9.84& 11.59& 17.97& 22.22& 31.80& 43.53 \\ 
 \hline
\end{tabular}
}
\caption{MSE for 12-layer, 8-head Transformer models (12L8AH) trained on degree-1 polynomials. During training, both input samples and coefficients are drawn from $\mathcal{U}(-1,1)$. At test time, input samples remain drawn from $\mathcal{U}(-1,1)$, while coefficients are drawn from $\mathcal{U}(-\sigma,\sigma)$. }
\label{table:training}
\end{table}

\subsection{Increasing Training Dataset Size}

Since dense coverage of the training distribution yields the strongest in-distribution performance, we next ask whether increasing the effective size of the training dataset would further improve accuracy. We keep the architecture and optimization procedure fixed while progressively increasing the number of training steps, with fresh samples drawn from the same training distribution at each step. Additional training steps therefore expose the model to a larger number of distinct training examples. We evaluate performance every 20k training steps.

\begin{figure}[!ht]
    \centering
    \begin{subfigure}{0.48\textwidth}
        \centering
        \includegraphics[width=\textwidth]{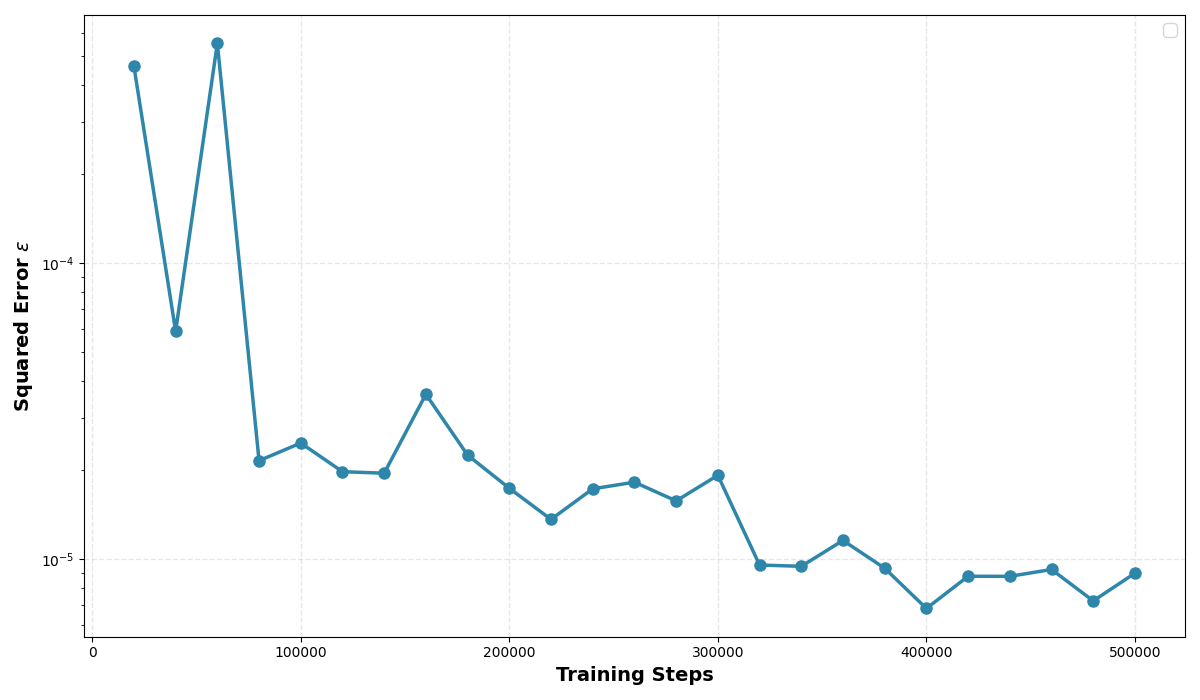}
        \caption{From step 0 (evaluated every 20K steps).}
        \label{fig:zigzag0}
    \end{subfigure}
    \hfill
    \begin{subfigure}{0.48\textwidth}
        \centering
        \includegraphics[width=\textwidth]{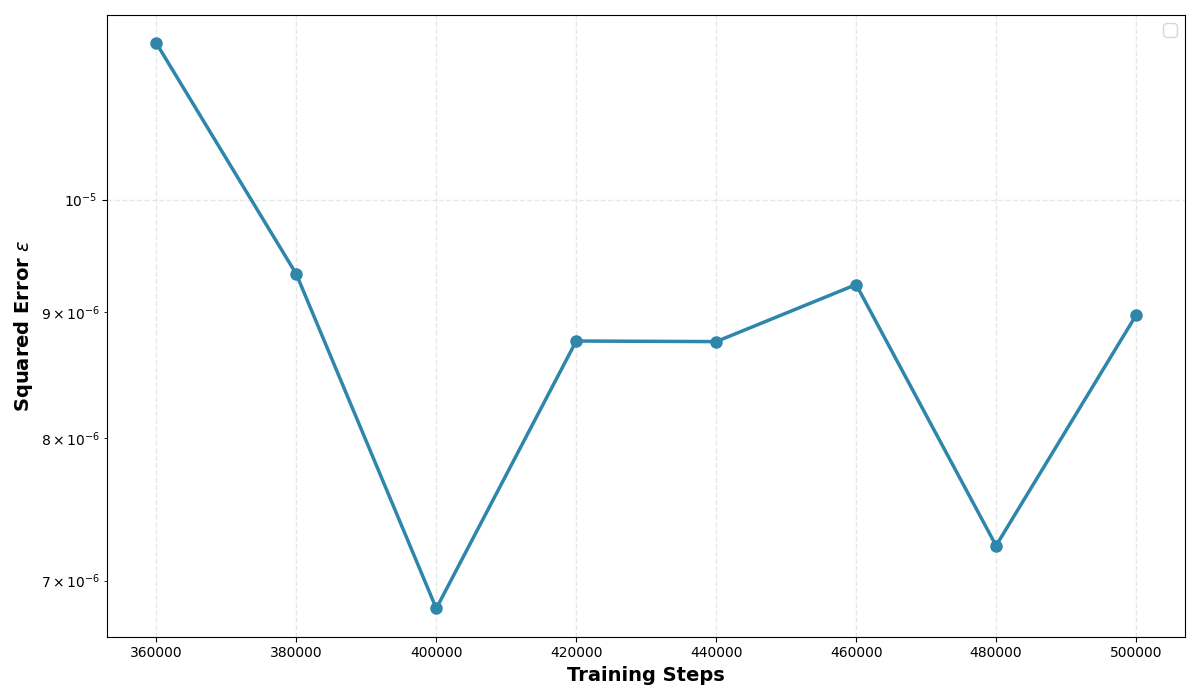}
        \caption{Zoom from step 360{,}000 (evaluated every 20K steps).}
        \label{fig:zigzag350000}
    \end{subfigure}
    \caption{In-distribution MSE during training for the degree-1 Transformer, evaluated every 20,000 optimization steps. Inputs and coefficients are sampled from $\mathcal{U}(-1,1)$ during both training and evaluation. Left: the full training trajectory. Right: a closer view from step 360,000 onward. Results are averaged over five evaluation seeds, with 100 test functions per seed, following Appendix~\ref{sec:trainingdetails}.}
    \label{fig:zigzag}
\end{figure}

Performance improves initially as training proceeds, but eventually plateaus, with evaluation error fluctuating within a narrow low-error range (Figure~\ref{fig:zigzag}). Thus, simply exposing the same architecture and optimization procedure to more freshly sampled training prompts does not further improve in-distribution accuracy, and longer training alone is unlikely to change the out-of-distribution behavior reported.

%does not resolve the generalization failure observed under distribution shift.

\label{appendix:shift}

\subsection{Expanding the training Distribution}

We next ask whether broader training support can improve robustness by reducing the extent of the distribution shift at test time. We train otherwise identical Transformers with both inputs and coefficients sampled from $\mathcal{U}(-a,a)$, for $a\in\{1,5,10,100\}$.

Moderately expanding the training distribution substantially improves robustness: models trained on $\mathcal{U}(-5,5)$ and $\mathcal{U}(-10,10)$ remain accurate over a much wider range of test scales than the baseline trained on $\mathcal{U}(-1,1)$. However, simply expanding the support is not a universal solution. Training on $\mathcal{U}(-100,100)$ under the same optimization procedure leads to large errors even for test scales contained within the training distribution (Table~\ref{table:7}).

Because this extreme setting also greatly increases the scale and variance of the targets, we do not interpret its failure as evidence of an architectural OOD limitation. Rather, the experiment shows that broader training support can delay the degradation, but naive domain expansion alone does not eliminate the problem.

\begin{table}[!ht]
\centering
\resizebox{\linewidth}{!}{
\begin{tabular}{|l|l|l|l|l|l|l|l|l|l|l|}
 \hline
 %\rowcolor{orange!30}
 models \ $\backslash$ \ $\sigma$ & 1 & 2 & 3 & 4 & 5 & 6 & 7 & 8 & 9 & 10 \\ 
 \hline
 ${\cal U}(-1,1)$   & $9 \times 10^{-6}$& 0.04 & 0.62& 1.95& 4.29& 7.14& 10.36& 15.79& 24.48& 27.06 \\ \hline

 ${\cal U}(-5,5)$   & 0.01& 0.01& 0.02& 0.03& 0.03& 0.05& 0.12& 0.27& 0.75& 1.61 \\ \hline

 ${\cal U}(-10,10)$   &0.13& 0.15& 0.17& 0.2& 0.26& 0.26& 0.32& 0.35& 0.41& 0.49 \\ \hline
 
  ${\cal U}(-100,100)$   &  2217.84& 2373.82& 2494.31& 2526.93& 2472.45& 2467.52& 2317.92& 2232.03& 2129.0& 2092.81 \\ %[1ex] 
 \hline
\end{tabular}
}

\caption{Effect of training-support width on MSE for degree-1 regression. Each model is trained with inputs and coefficients sampled from $\mathcal{U}(-a,a)$, where $a\in\{1,5,10,100\}$. All models are evaluated with inputs sampled from $\mathcal{U}(-1,1)$ and coefficients sampled from $\mathcal{U}(-\sigma,\sigma)$.}
\label{table:7}
\end{table}

\subsection{Degree-Schedule Interventions}
\label{sec:extrapolation}

\begin{figure*}[!ht]
\includegraphics[width=\textwidth, height=5.9cm]{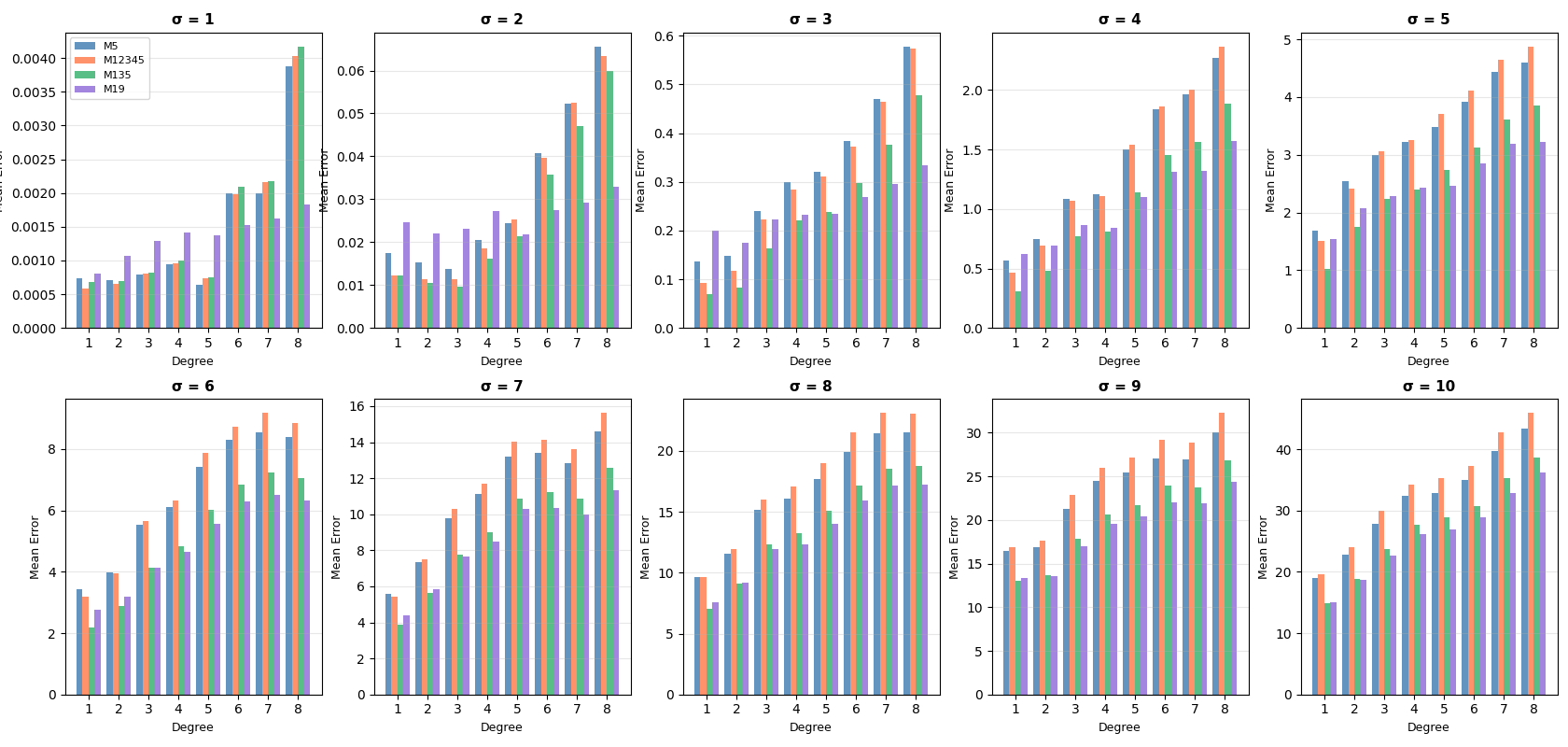}
\caption{Bar charts showing the evolution of the mean squared error $\epsilon$ for Transformer models with 12 layers and 8 attention heads across four training strategies, tested on degrees 1 to 8 to evaluate performance across trained, intermediate, and unseen degrees. M5 is trained only on degree 5, M12345 on degrees 1--5, M135 on degrees 1, 3, and 5, and M19 on degrees 1 and 9 (black edges indicate training degrees). During training, both input samples and coefficients are drawn from $\mathcal{U}(-1,1)$. At test time, input samples remain drawn from $\mathcal{U}(-1,1)$, while coefficients are drawn from $\mathcal{U}(-\sigma,\sigma)$ with $\sigma \in \{1,\ldots,10\}$ to assess robustness to out-of-distribution coefficient magnitudes. Results are averaged over five random seeds (42, 123, 456, 789, 1011), with each seed evaluated on 100 functions, following the evaluation protocol described in Appendix \ref{sec:trainingdetails}. Note that each plot uses its own scale.}
\label{fig:ood_performance}
\end{figure*}

To test whether the structure of degree exposure during training affects robustness under distribution shift, we compare: 
\begin{enumerate}
    \item \textbf{Single-degree training} trains exclusively on polynomials of a fixed degree $n$.
    
    \item \textbf{Curriculum training} trains on polynomial classes in increasing degree order, from degree 1 to $n$.

    \item \textbf{Sparse-degree training} trains on non-consecutive polynomial degree classes, skipping intermediate ones (e.g., degrees 1, 3, and 5).
   
    \item \textbf{Extreme-gap training} trains only on low- and high-degree polynomials, leaving a large gap of unseen intermediate degrees (e.g., degrees 1 and 9).
\end{enumerate}

\begin{figure}[!ht]
    \centering
    \includegraphics[width=\linewidth, height = 4.3cm]{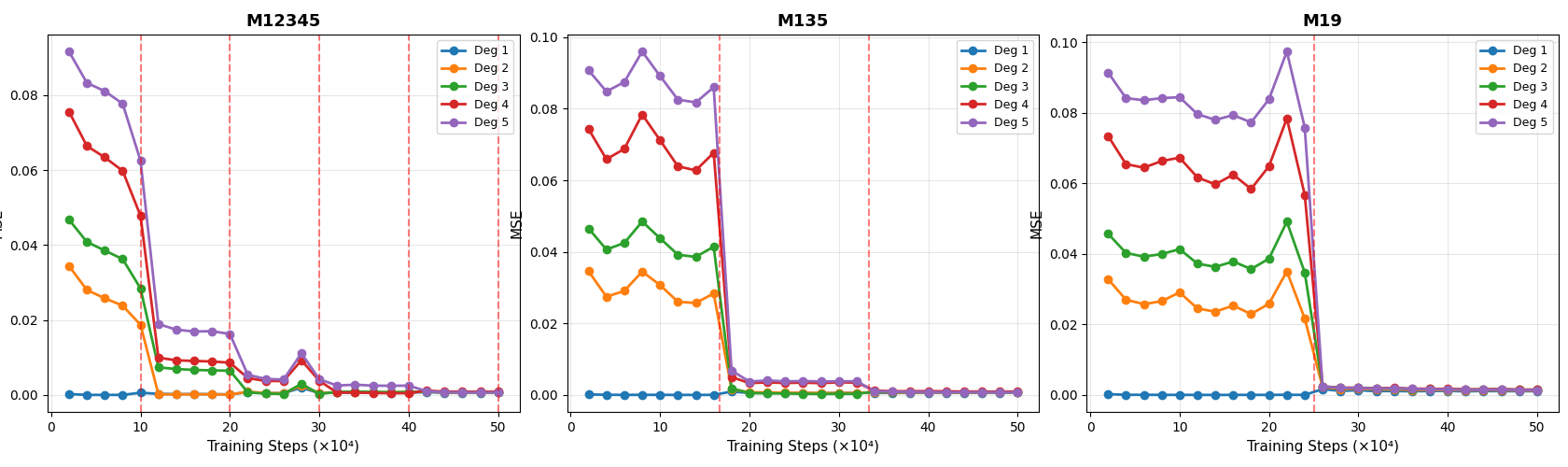}
    \caption{Evolution of the MSE during training, evaluated every 20k steps on intermediate degrees (1–5). Model M12345 was trained sequentially on degrees 1–5 for 100k steps per degree (curriculum); M135 on degrees 1, 3, and 5 for 166.7k steps per degree (sparse-degree); and M19 on degrees 1 and 9 for 250k steps per degree (extreme-gap). Vertical dashed lines indicate transitions between curriculum stages. }
    \label{generalization_dynamics}
\end{figure}

These experiments are intended as training-schedule interventions: they test whether changing the sequence of polynomial degrees seen during training can remove the distribution-shift failure observed in the main experiments. Because the multi-degree models are trained sequentially rather than on a joint mixture of degrees, their final performance can reflect both transfer between degrees and catastrophic forgetting of earlier degrees. This effect is visible in Figure~\ref{generalization_dynamics}, where the error on earlier degrees can increase after the training task switches to later degrees. Therefore, these experiments should not be interpreted as a clean isolation of the model's intrinsic ability to interpolate or extrapolate across polynomial degrees.

\begin{figure*}[!h] 
\includegraphics[width=\textwidth]{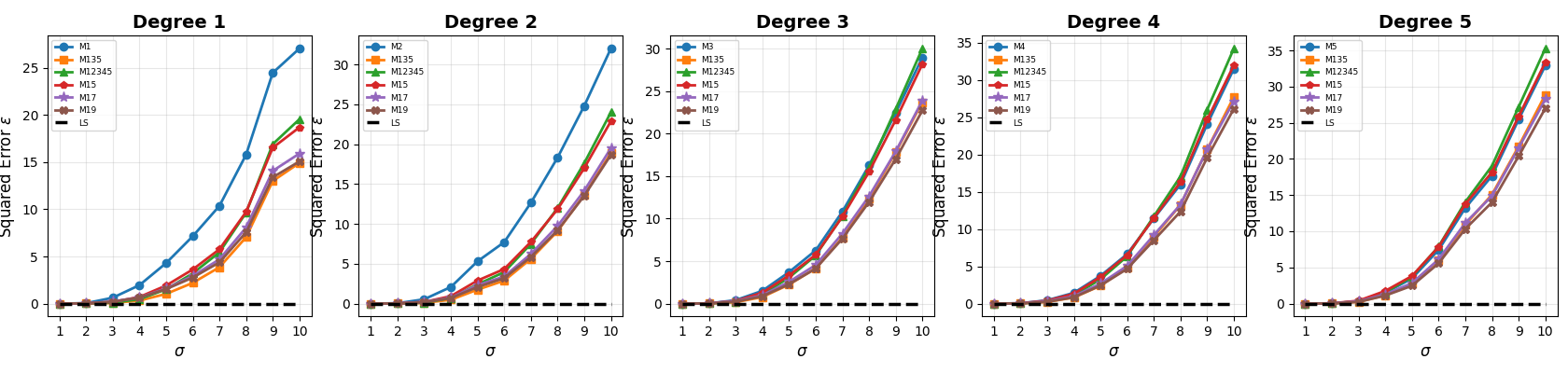}
\caption{Evolution of the MSE for Transformer models (12 layers, 8 attention heads). Models are trained from scratch on different polynomial degrees: $M_i$ is trained on degree $i$ only, while $M_{15}$, $M_{17}$, $M_{19}$, $M_{135}$, and $M_{12345}$ denote training on degrees $\{1,5\}$, $\{1,7\}$, $\{1,9\}$, $\{1,3,5\}$, and $\{1,\dots,5\}$, respectively. During training, inputs and coefficients are sampled from $\mathcal{U}(-1,1)$. At test time, inputs remain in $\mathcal{U}(-1,1)$, while coefficients are sampled from $\mathcal{U}(-\sigma,\sigma)$ with $\sigma \in \{1,\dots,10\}$. All models are evaluated on degrees 1--5. 
}
\label{generalization}
\end{figure*}

Figures~\ref{fig:ood_performance} and~\ref{generalization} show that the sparse-degree and extreme-gap schedules can improve performance on some intermediate degrees under distribution shift, while maintaining comparable in-distribution performance in several cases. However, because of the sequential training protocol, these gains may reflect a combination of transfer, optimization path dependence, and forgetting, rather than pure degree interpolation. In contrast, extrapolation to higher unseen degrees remains challenging across strategies.

Overall, changing the degree-training schedule affects the quantitative error profile, but it does not remove the qualitative distribution-shift failure. We therefore use these experiments only as evidence that this class of training-schedule interventions does not by itself solve the generalization problem; a joint-mixture training protocol would be required to isolate degree interpolation from catastrophic forgetting.

%\newpage

\newpage

\subsection{Gradual input domain shift}

\paragraph{Limits to Extrapolation Across Sequence Intervals}

\begin{figure}[!h]
    \centering
    \begin{subfigure}[b]{0.32\textwidth}
        \centering
        \includegraphics[width=\textwidth]{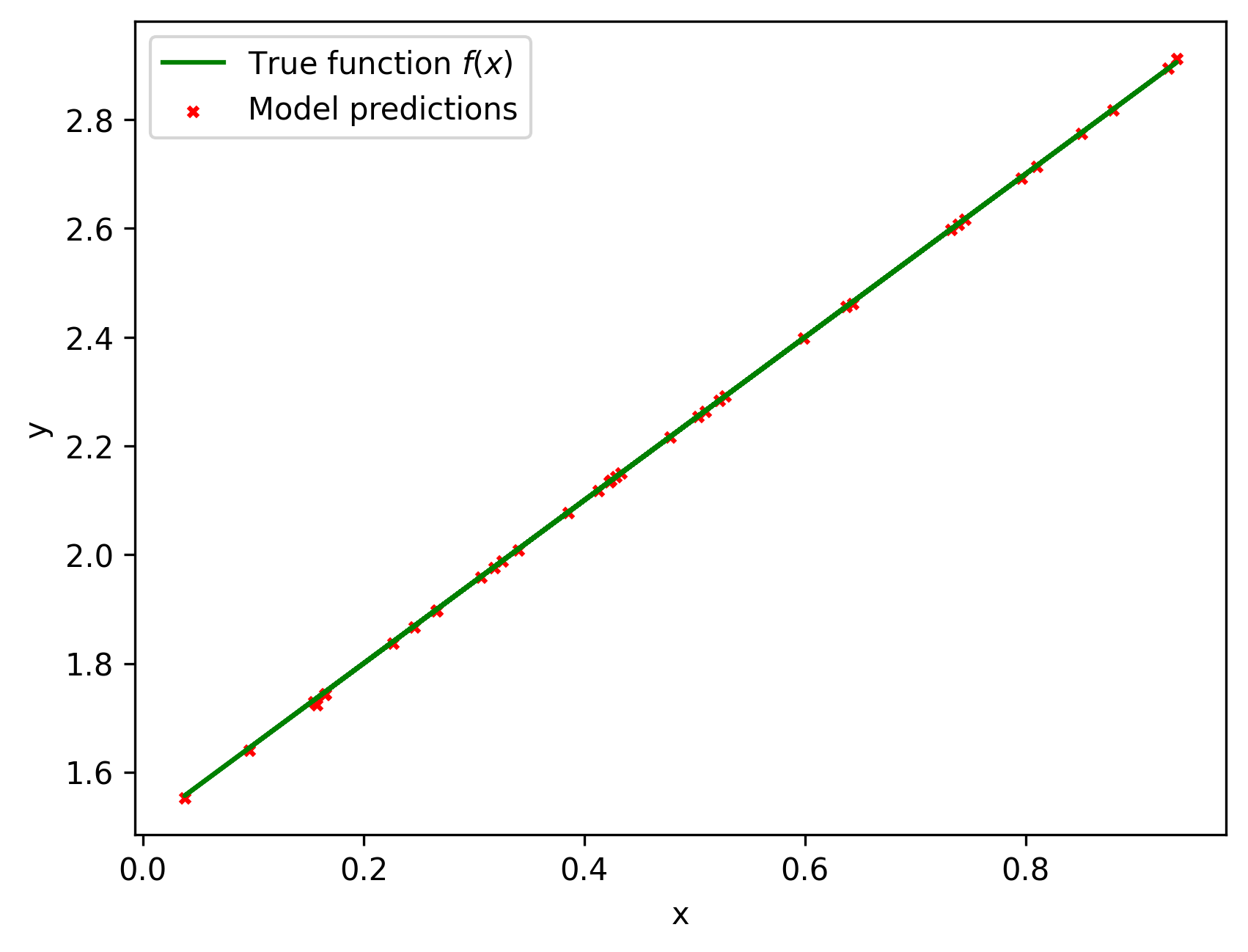}
        \caption{Scenario 1: $x \in [0,1]$}
        \label{fig:scenario1}
    \end{subfigure}
    \hfill
    \begin{subfigure}[b]{0.32\textwidth}
        \centering
        \includegraphics[width=\textwidth]{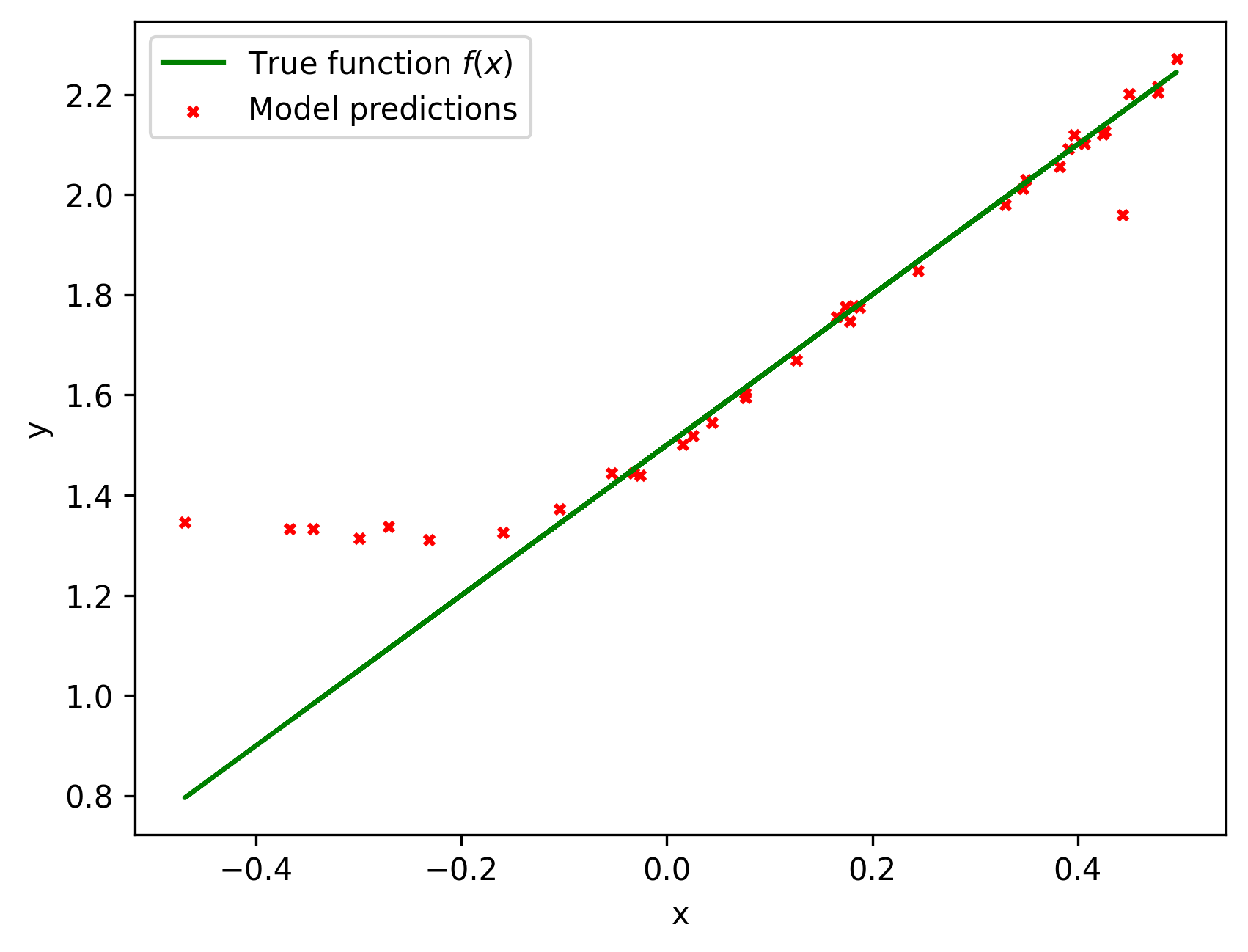}
        \caption{Scenario 2: $x \in [-0.5,0.5]$}
        \label{fig:scenario2}
    \end{subfigure}
    \hfill
    \begin{subfigure}[b]{0.32\textwidth}
        \centering
        \includegraphics[width=\textwidth]{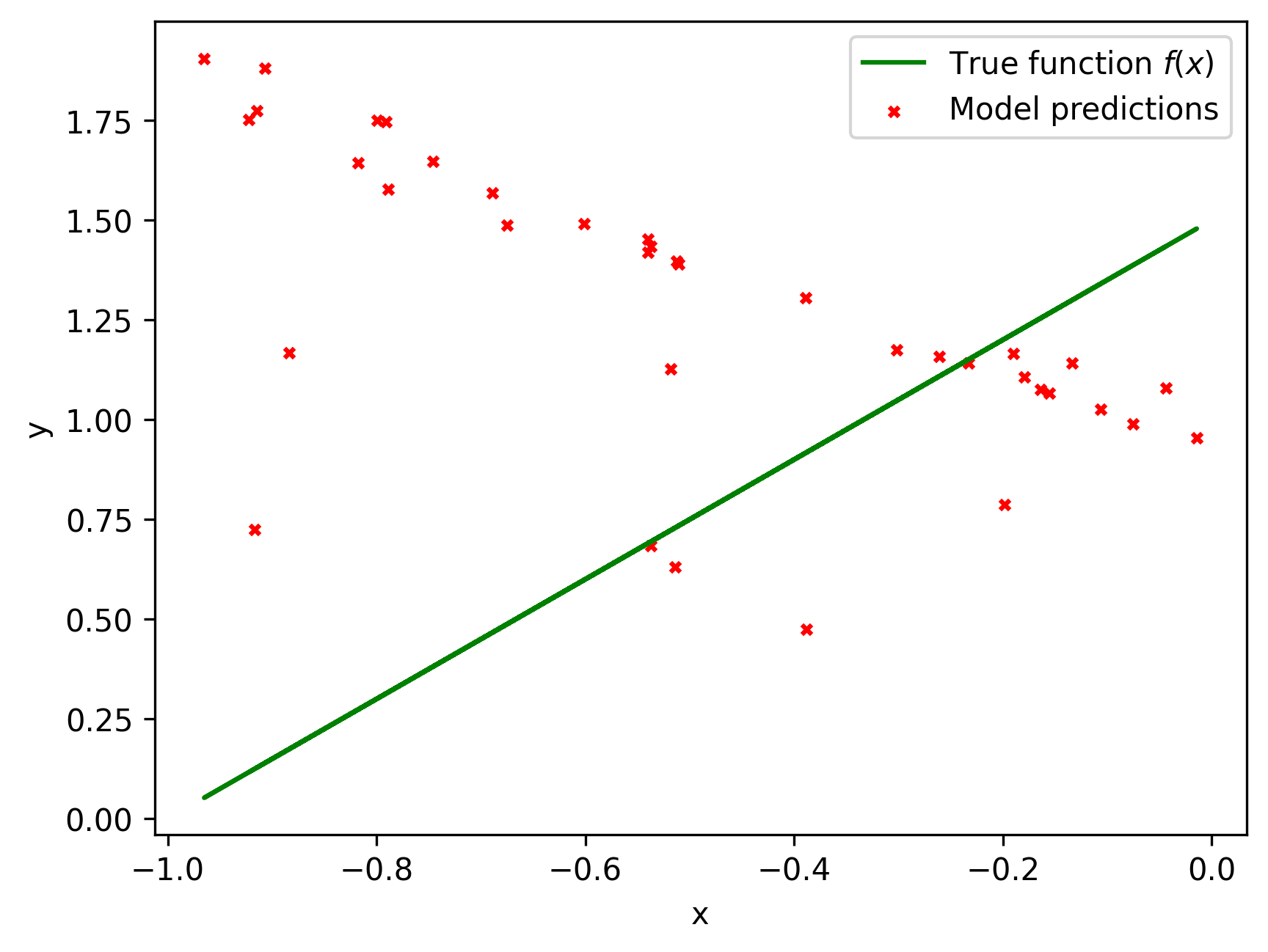}
        \caption{Scenario 3: $x \in [-1,0]$}
        \label{fig:scenario3}
    \end{subfigure}
    
   % \vspace{0.5em}
    
    \begin{subfigure}[b]{0.32\textwidth}
        \centering
        \includegraphics[width=\textwidth]{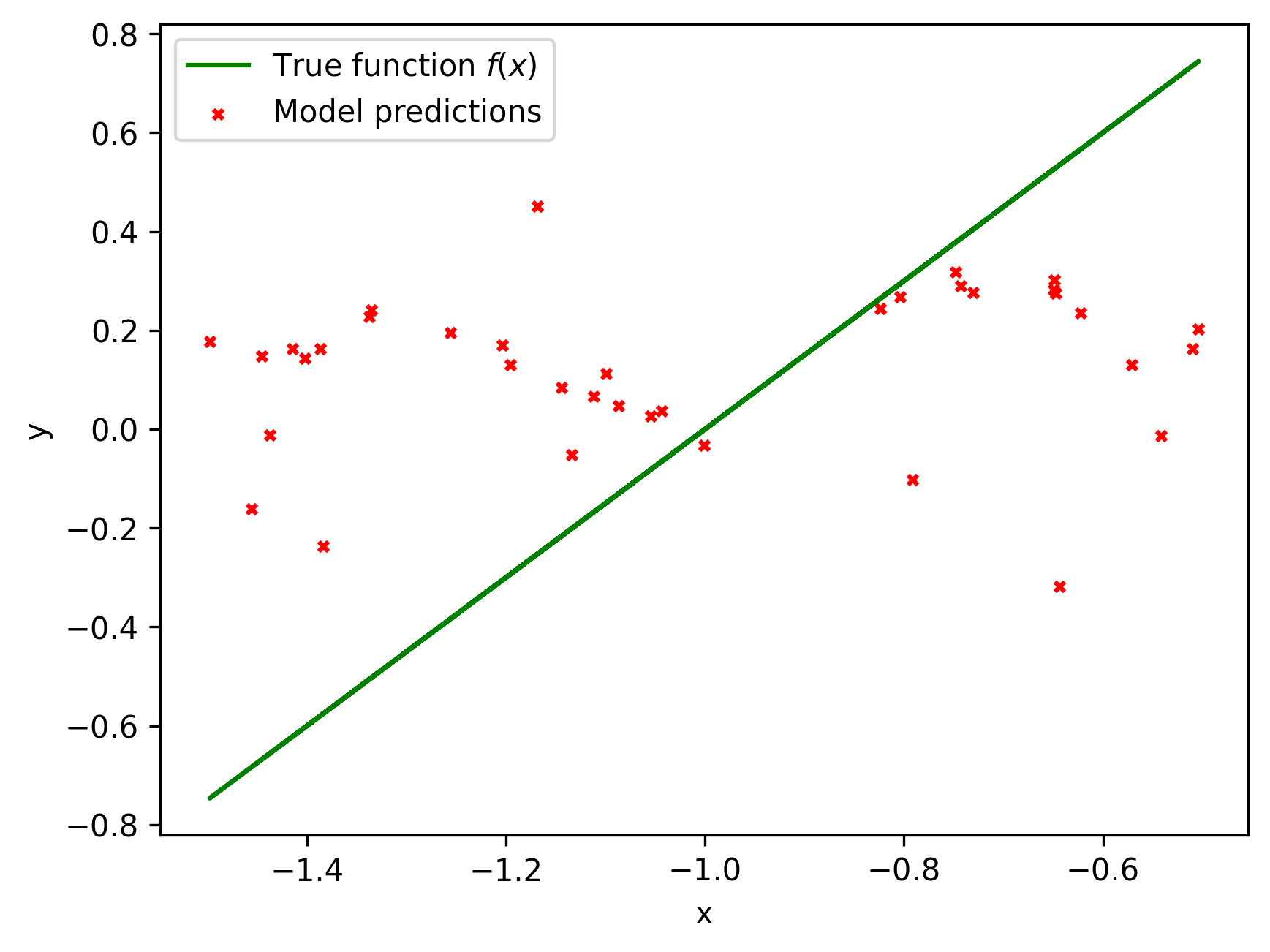}
        \caption{Scenario 4: $x \in [-1.5,-0.5]$}
        \label{fig:scenario4}
    \end{subfigure}
    \hfill
    \begin{subfigure}[b]{0.32\textwidth}
        \centering
        \includegraphics[width=\textwidth]{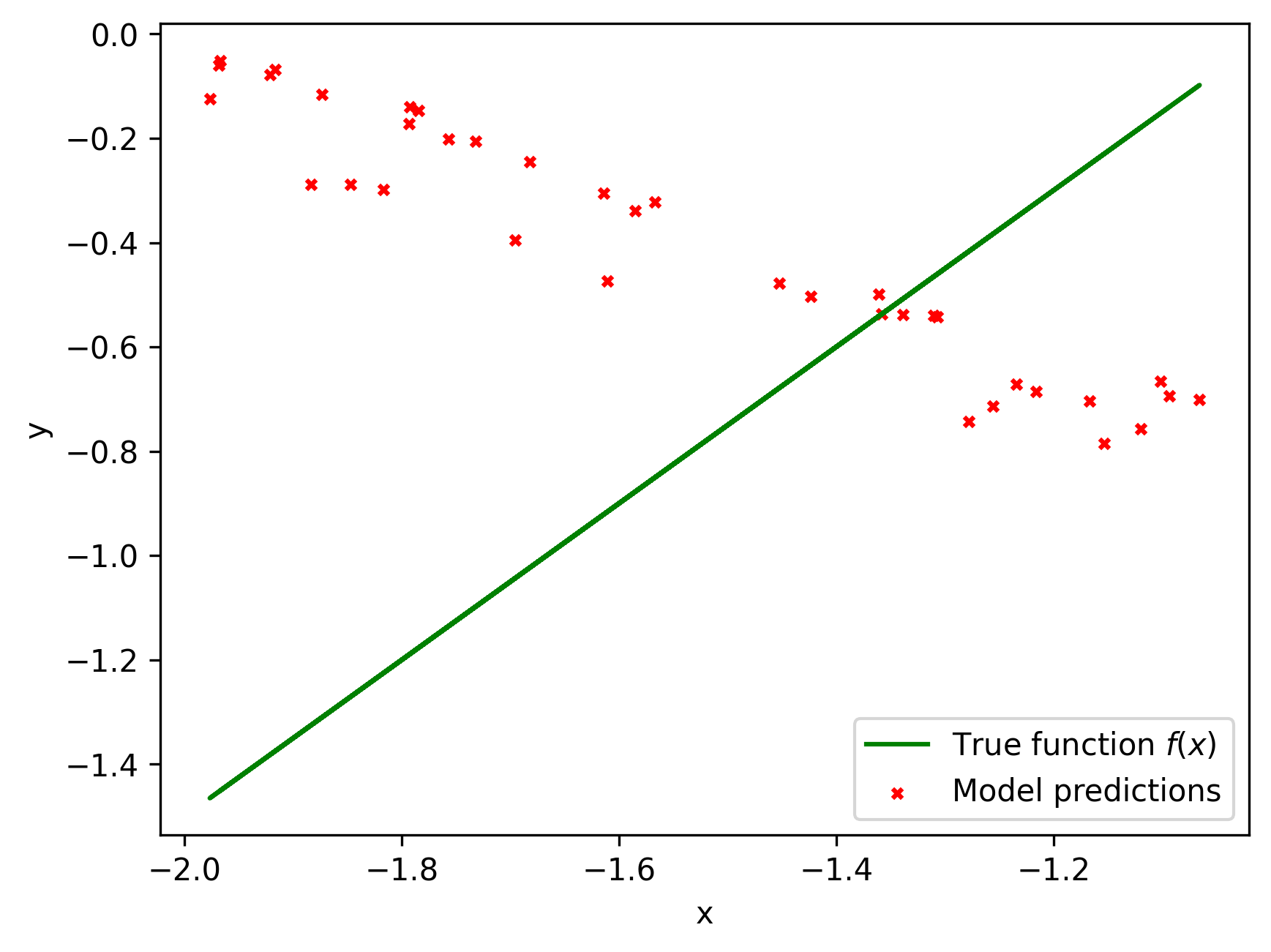}
        \caption{Scenario 5: $x \in [-2,-1]$}
        \label{fig:scenario5}
    \end{subfigure}
    \hfill
    \begin{subfigure}[b]{0.32\textwidth}
        \centering
        \includegraphics[width=\textwidth]{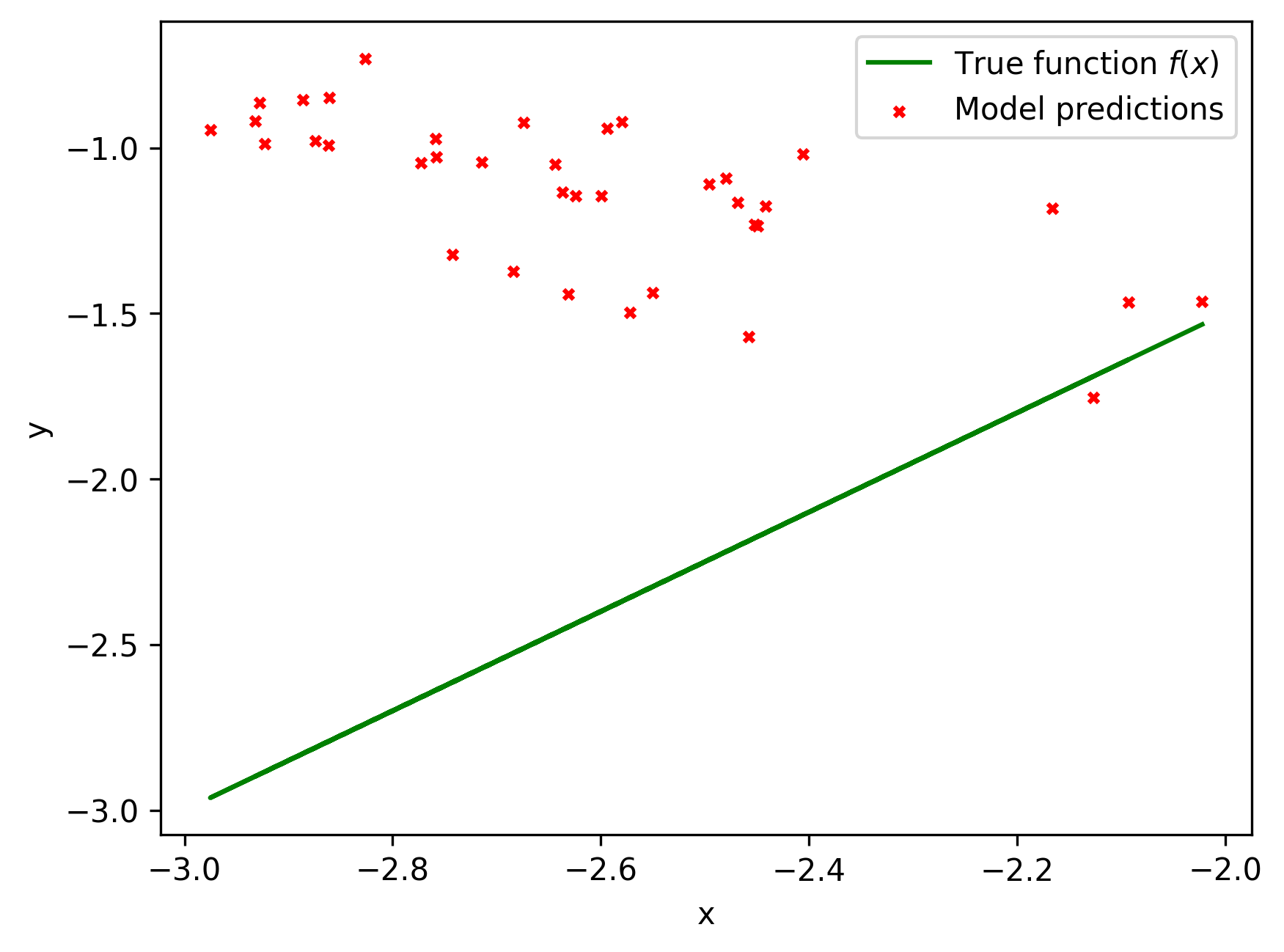}
        \caption{Scenario 6: $x \in [-3,-2]$}
        \label{fig:scenario6}
    \end{subfigure}
    
\caption{Gradual domain shift for f(x) = 1.5x+ 1.5, progressively entering the unseen region.
The model was trained on linear functions $f(x) = ax+ b$ with inputs $x \in [0,1]$ and coefficients $a,b \in [-1,0] \cup [1,2]$}

    \label{intervals}
\end{figure}

To further probe the transformer's generalization behavior, we examine how it extrapolates to sequences outside the regions encountered during training. We train models on linear functions \(f(x) = ax + b\) with \(x \in [0,1]\) and coefficients \(a, b \in [-1,0] \cup [1,2]\), creating a gap in the range of values observed during training. At evaluation time, we consider the function $f(x) = 1.5x + 1.5$ and gradually shift the input interval toward regions that were never observed during training. 
While the model performs accurately on sequences within familiar intervals, its predictions fail abruptly when the input enters an unseen region, even though the underlying function remains linear (Figure \ref{intervals}). This failure suggests that the model primarily relies on recognizing sequences similar to those encountered during training rather than learning a general representation.%, and fails dramatically when presented with sequences outside these familiar patterns.

\newpage
\paragraph{Support shift without magnitude extrapolation.} To separate distribution shift from magnitude extrapolation, we train a degree-1 Transformer on $x\in[0,1]$ and evaluate it on the disjoint interval $[-1,0]$, which preserves the same range of input magnitudes. The model is accurate on the training support but degrades sharply once inputs enter the unseen interval. Thus, the cliff does not require larger input magnitudes; a change in input support alone is sufficient to induce the failure.

\begin{figure}[!h]
    \centering
    \begin{subfigure}[b]{0.32\textwidth}
        \centering
        \includegraphics[width=\textwidth]{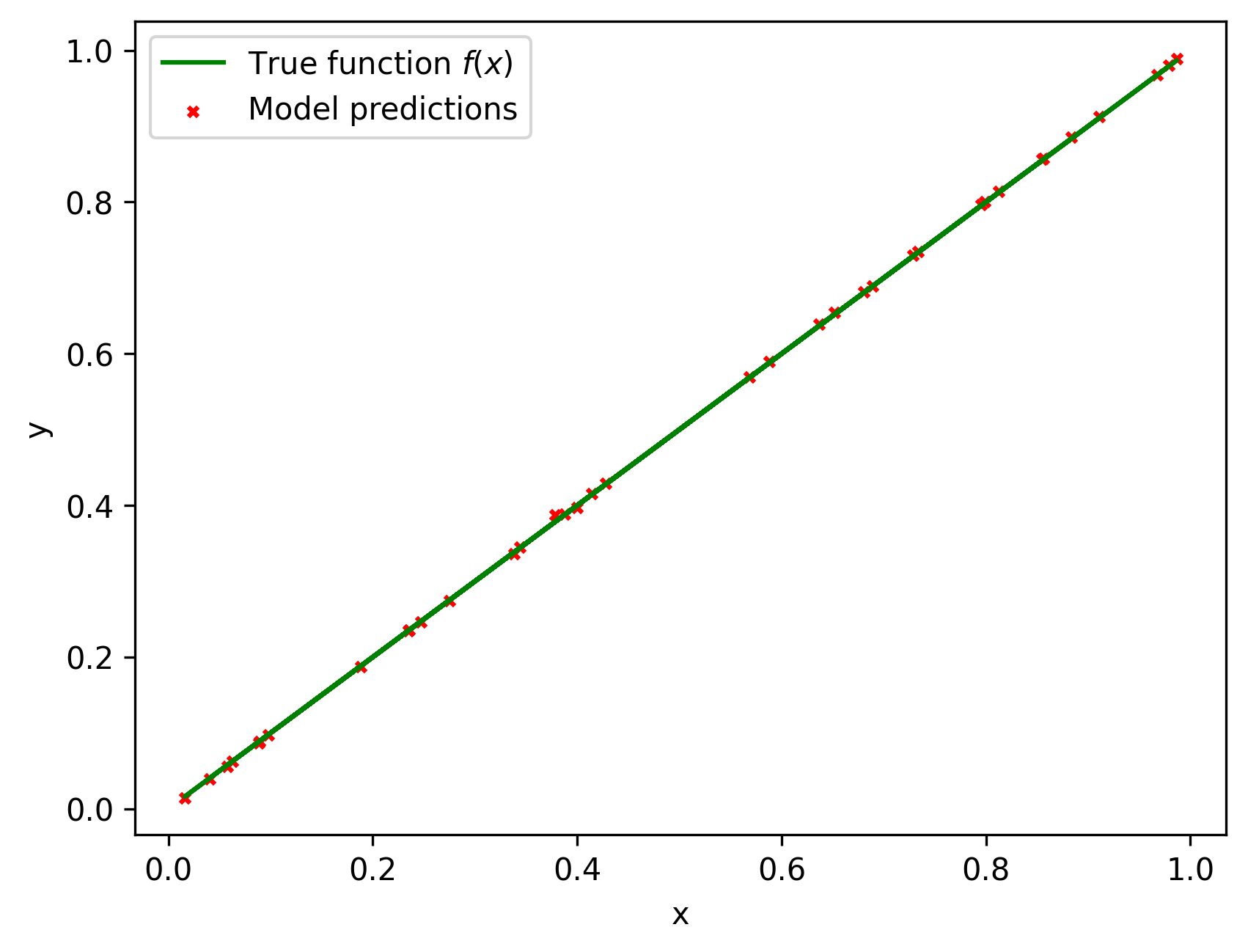}
        \caption{Scenario 1: $x \in [0,1]$}
        \label{fig:scenario1}
    \end{subfigure}
    \hfill
    \begin{subfigure}[b]{0.32\textwidth}
        \centering
        \includegraphics[width=\textwidth]{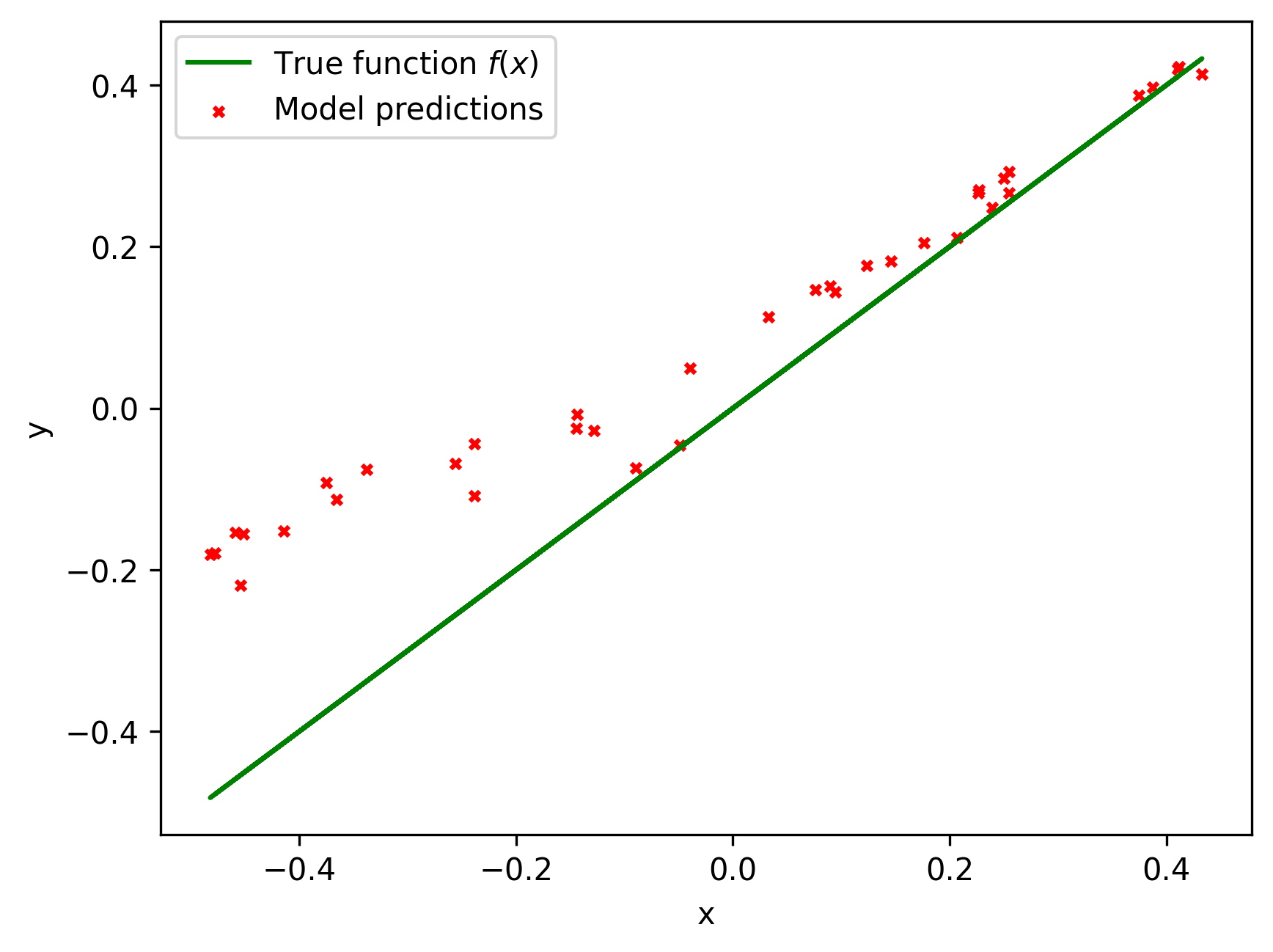}
        \caption{Scenario 2: $x \in [-0.5,0.5]$}
        \label{fig:scenario2}
    \end{subfigure}
    \hfill
    \begin{subfigure}[b]{0.32\textwidth}
        \centering
        \includegraphics[width=\textwidth]{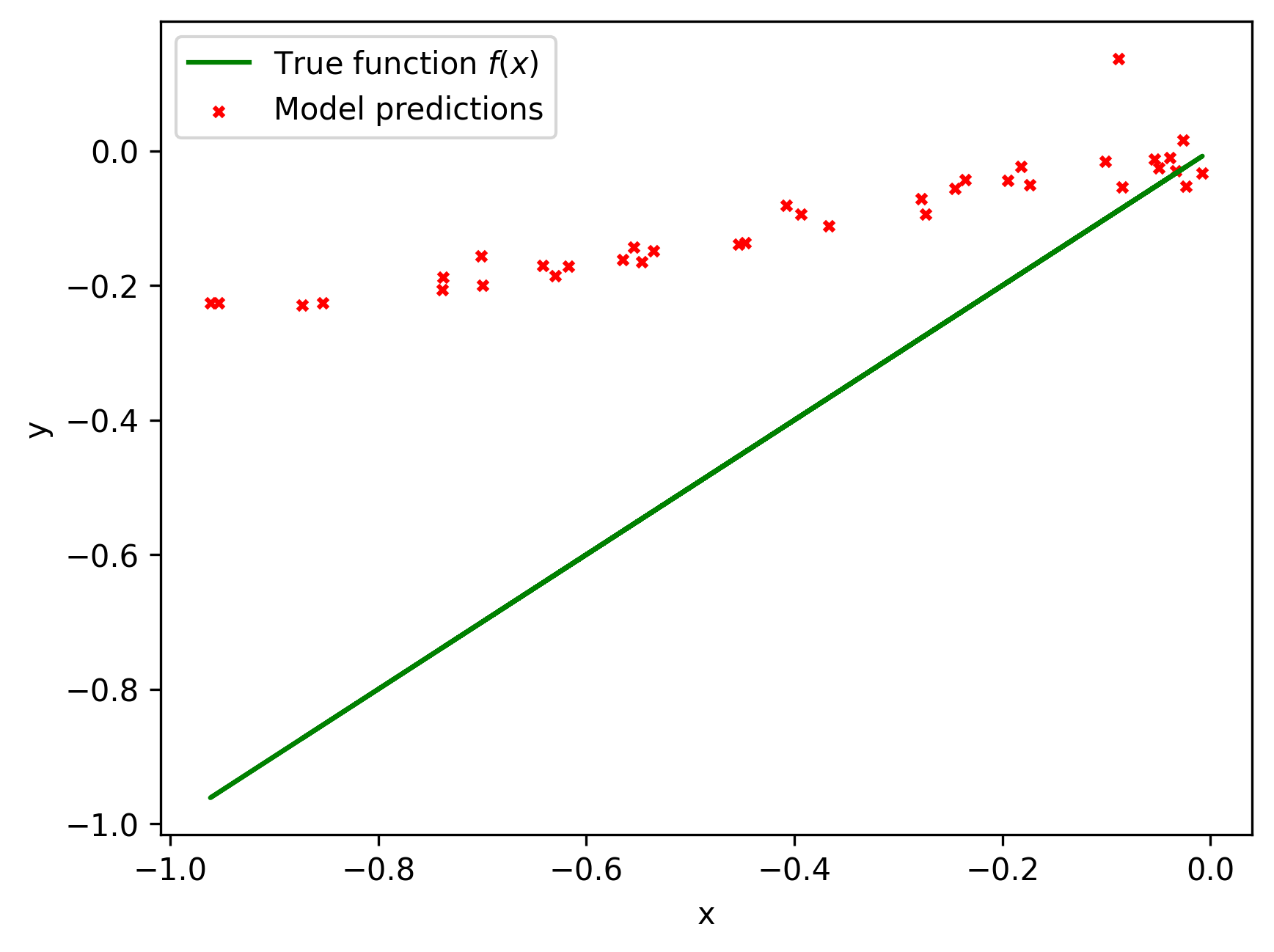}
        \caption{Scenario 3: $x \in [-1,0]$}
        \label{fig:scenario3}
    \end{subfigure}

\caption{Input-support shift at fixed magnitude. The model is trained on inputs $x\in[0,1]$ and evaluated on the disjoint interval $[-1,0]$, which preserves the same input-magnitude range. The resulting degradation shows that the failure is not explained by magnitude extrapolation alone.}

    \label{intervals_inputs}
\end{figure}

%\newpage

\subsection{Nonlinear Encoder and Decoder Mappings}
\label{app:mlp_encoding}

To test whether the observed failure is a consequence of the simple linear mappings used to encode inputs and decode predictions, we replace both the standard linear encoder and decoder with nonlinear MLPs while keeping the remainder of the architecture and training procedure unchanged. 

We evaluate the resulting models under the same distribution shift as in the main experiments: inputs remain sampled from $\mathcal{U}(-1,1)$, while polynomial coefficients are sampled from $\mathcal{U}(-\sigma,\sigma)$ for $\sigma\in\{1,\ldots,10\}$. Tables~\ref{tab:mlp_mse} and~\ref{tab:mlp_nmse} report the corresponding MSE and NMSE results.

\begin{table}[!ht]
\centering
\resizebox{\textwidth}{!}{
\begin{tabular}{lcccccccccc}
\toprule
Degree
& $\sigma=1$ & $\sigma=2$ & $\sigma=3$ & $\sigma=4$ & $\sigma=5$
& $\sigma=6$ & $\sigma=7$ & $\sigma=8$ & $\sigma=9$ & $\sigma=10$ \\
\midrule

1
& $(9.5\pm0.8)\times10^{-6}$
& $(4.5\pm0.8)\times10^{-2}$
& $0.6\pm0.0$
& $2.1\pm0.3$
& $4.4\pm0.4$
& $6.9\pm0.6$
& $12.3\pm1.6$
& $16.5\pm1.6$
& $22.3\pm1.4$
& $30.9\pm1.5$ \\

2
& $(2.8\pm0.2)\times10^{-5}$
& $(2.7\pm1.0)\times10^{-2}$
& $0.4\pm0.0$
& $1.6\pm0.1$
& $3.6\pm0.2$
& $6.9\pm0.4$
& $11.9\pm1.4$
& $15.7\pm2.0$
& $23.1\pm1.9$
& $30.0\pm3.0$ \\

3
& $(1.0\pm0.1)\times10^{-4}$
& $(2.4\pm0.6)\times10^{-2}$
& $0.3\pm0.0$
& $1.4\pm0.2$
& $3.3\pm0.3$
& $6.3\pm0.6$
& $10.9\pm1.2$
& $15.6\pm1.5$
& $23.4\pm2.0$
& $29.4\pm2.7$ \\

4
& $(7.3\pm0.3)\times10^{-4}$
& $(1.7\pm0.3)\times10^{-2}$
& $0.2\pm0.0$
& $1.0\pm0.2$
& $2.8\pm0.3$
& $5.8\pm0.8$
& $10.1\pm1.5$
& $14.2\pm0.8$
& $21.4\pm1.3$
& $29.9\pm3.5$ \\

5
& $(1.4\pm0.1)\times10^{-3}$
& $(1.8\pm0.2)\times10^{-2}$
& $0.2\pm0.0$
& $0.8\pm0.1$
& $2.4\pm0.2$
& $5.1\pm0.5$
& $8.6\pm1.1$
& $12.6\pm1.3$
& $19.6\pm0.3$
& $28.2\pm1.2$ \\

6
& $(1.5\pm0.1)\times10^{-3}$
& $(1.4\pm0.1)\times10^{-2}$
& $0.2\pm0.0$
& $0.8\pm0.2$
& $2.6\pm0.4$
& $5.5\pm0.4$
& $9.4\pm1.1$
& $14.6\pm1.1$
& $22.3\pm1.2$
& $30.0\pm1.2$ \\

\bottomrule
\end{tabular}
}
\caption{Evolution of the MSE for Transformer models with 12 layers and 8 attention heads, using nonlinear MLP encoder and decoder, trained on polynomials of degree $i \in \{1,\dots,6\}$. During training, both input samples and coefficients are drawn from $\mathcal{U}(-1,1)$. At test time, input samples remain drawn from $\mathcal{U}(-1,1)$, while coefficients are drawn from $\mathcal{U}(-\sigma,\sigma)$ with $\sigma \in \{1,\cdots,10\}$. Each model is evaluated on its corresponding degree. Results are averaged over five random seeds (42, 123, 456, 789, 1011), with each seed evaluated on 100 functions, following the evaluation protocol described in Appendix \ref{sec:trainingdetails}.}
\label{tab:mlp_mse}
\end{table}

\begin{table}[!ht]
\centering
\resizebox{\textwidth}{!}{
\begin{tabular}{lcccccccccc}
\toprule
Degree
& $\sigma=1$ & $\sigma=2$ & $\sigma=3$ & $\sigma=4$ & $\sigma=5$
& $\sigma=6$ & $\sigma=7$ & $\sigma=8$ & $\sigma=9$ & $\sigma=10$ \\
\midrule
1
& $2.19{\times}10^{-5}$ & $0.025$ & $0.154$ & $0.287$ & $0.393$
& $0.465$ & $0.548$ & $0.591$ & $0.640$ & $0.678$ \\

2
& $5.44{\times}10^{-5}$ & $0.013$ & $0.085$ & $0.198$ & $0.296$
& $0.383$ & $0.463$ & $0.502$ & $0.561$ & $0.602$ \\

3
& $1.84{\times}10^{-4}$ & $0.011$ & $0.066$ & $0.161$ & $0.241$
& $0.322$ & $0.401$ & $0.455$ & $0.510$ & $0.555$ \\

4
& $0.001$ & $0.007$ & $0.044$ & $0.104$ & $0.191$
& $0.279$ & $0.349$ & $0.398$ & $0.460$ & $0.512$ \\

5
& $0.002$ & $0.008$ & $0.033$ & $0.085$ & $0.151$
& $0.221$ & $0.286$ & $0.331$ & $0.398$ & $0.456$ \\

6
& $0.002$ & $0.006$ & $0.028$ & $0.084$ & $0.158$
& $0.235$ & $0.301$ & $0.361$ & $0.432$ & $0.475$ \\
\bottomrule
\end{tabular}
}
\caption{NMSE for the Transformers with nonlinear MLP encoders and decoders evaluated in Table~\ref{tab:mlp_mse}. Each model is evaluated on its training degree under the same coefficient-scaling protocol. Entries are means over five evaluation seeds.}
\label{tab:mlp_nmse}
\end{table}

\newpage

The nonlinear encoder and decoder mappings exhibit the same qualitative behavior as the linear mappings, including the same boundary value behavior (Table \ref{tab:mlp_boundaries}).

\begin{table}[!ht]
\centering
\begin{tabular}{ccc}
\toprule
Degree & $B_d^{+}$ & $B_d^{-}$ \\
\midrule
1 & $-0.389$ & $0.146$ \\
2 & $-0.483$ & $0.230$ \\
3 & $-0.350$ & $0.563$ \\
4 & $-1.134$ & $2.035$ \\
5 & $-2.385$ & $1.681$ \\
6 & $-0.460$ & $0.482$ \\
\bottomrule
\end{tabular}
\caption{Estimated boundary values for models with nonlinear MLP encoder and decoder mappings. $B_d^{+}$ denotes the value at which predictions saturate as the query takes large positive values, and $B_d^{-}$ the value at which they saturate as the query takes large negative values.}
\label{tab:mlp_boundaries}
\end{table}

\subsection{Prompt-wise Label Normalization}
\label{app:label_normalization}

Since coefficient scaling increases the magnitude of the labels, it may push their embeddings far outside the range seen during training. To test whether the cliff is simply an artifact of this fixed encoding, we normalize each prompt by its own label scale and rescale the prediction back afterward.

\begin{table}[!ht]
\centering
\small
\setlength{\tabcolsep}{4pt}
\label{tab:label_normalization}
\resizebox{\linewidth}{!}{
\begin{tabular}{lcccccccccc}
\toprule
 & $\sigma=1$ & $2$ & $3$ & $4$ & $5$ & $6$ & $7$ & $8$ & $9$ & $10$ \\
\midrule
Baseline
& $9 \times 10^{-6}$& 0.04 & 0.62& 1.95& 4.29& 7.14& 10.36& 15.79& 24.48& 27.06 \\ 

Label-normalized
& 0.10
& 0.45
& 0.99
& 1.76
& 2.72
& 3.60
& 5.49
& 7.09
& 8.76
& 11.32 \\
\bottomrule
\end{tabular}
}
\caption{Prompt-wise label normalization for degree-1 polynomial regression under coefficient-scale shift.}
\end{table}

Prompt-wise normalization improves performance in the far-OOD regime, but degrades in- and near-distribution accuracy. Thus, although the label scale contributes to the degradation, the cliff cannot be explained solely by the embeddings leaving their training range.

%\newpage

\section{Boundary Values persist across different training degrees}

\begin{figure}[!h] 
     \includegraphics[width=0.33\linewidth]{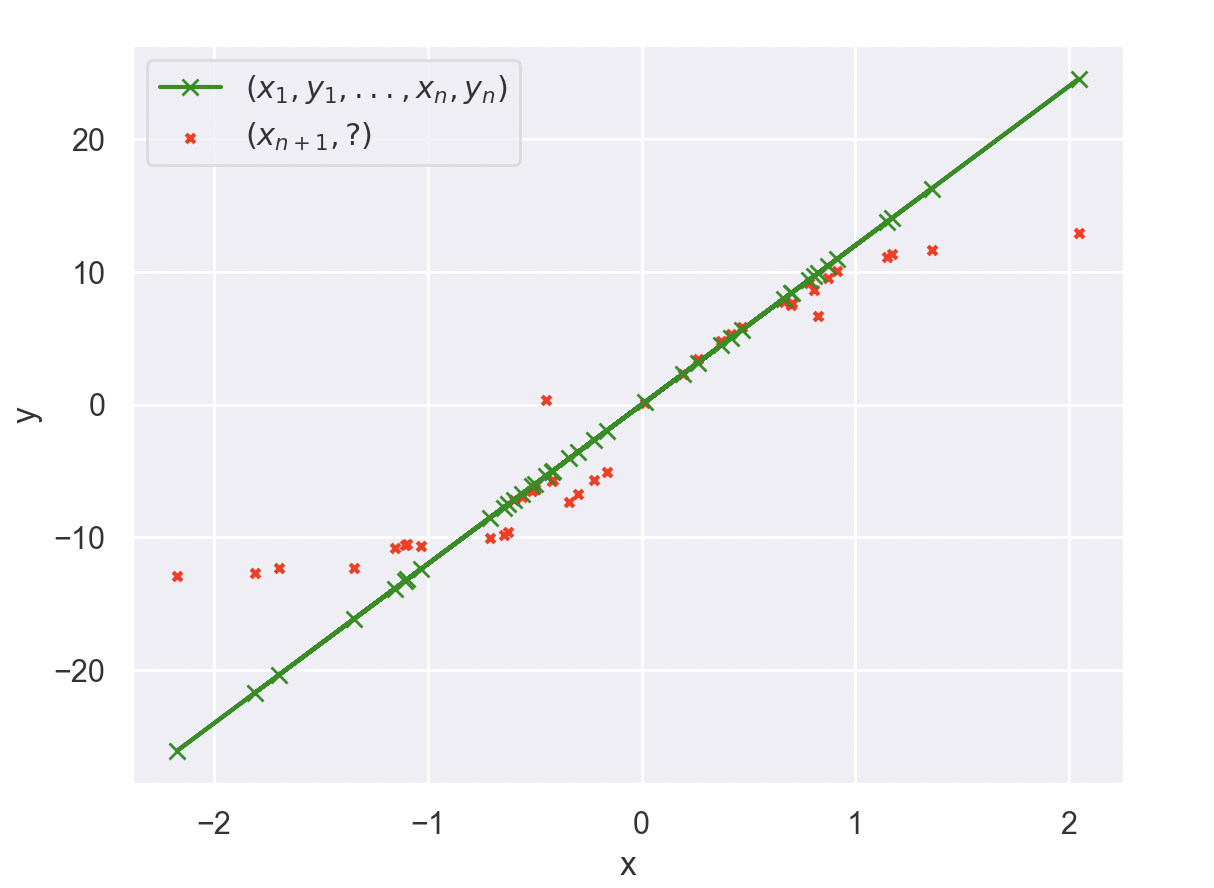}
    \includegraphics[width=0.33\linewidth]{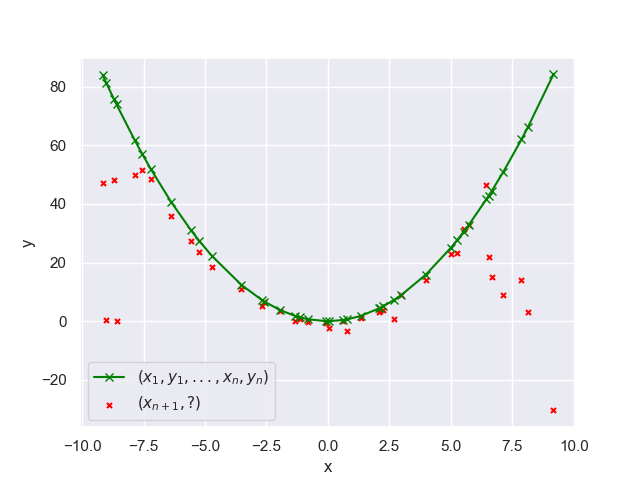}
    \includegraphics[width=0.33\linewidth]{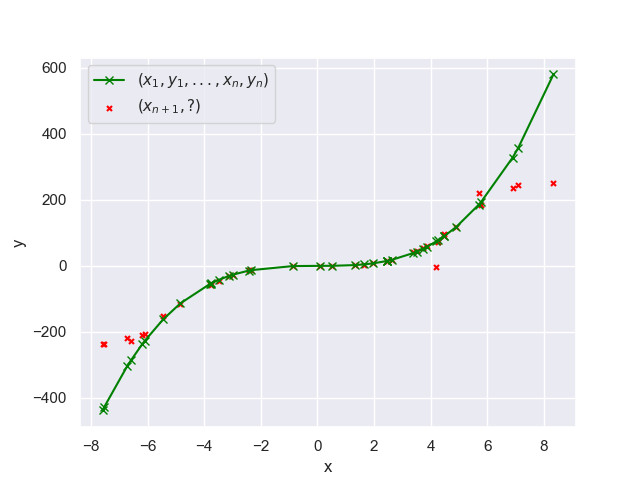} 
   
\caption{Emergence of boundary values in models trained and evaluated on polynomial functions. Results are presented for degree 1 (left), degree 2 (middle), and degree 3 (right). }\label{sequence1}
\end{figure}

%\newpage

\section{Linear Attention Can Reproduce a Gradient Descent Prediction}
\label{sec:theory}

We present a constructive result showing that a single-layer transformer with linear attention can exactly reproduce the \emph{prediction} obtained from one gradient descent step for linear regression. The result serves as a theoretical baseline illustrating how attention can emulate optimization dynamics under strong simplifying assumptions. These assumptions are violated in practical transformers, and the construction should not be interpreted as an explanation of empirical ICL behavior.

\subsection{Problem Setup}

\paragraph{Task.}
We consider prompts represented as a sequence of concatenated input-output tokens
\[
  e_i = (x_i,y_i), \qquad i=1,\ldots,n,
\]
followed by a query token
\[
  e_{n+1} = (x_{\mathrm{query}},0).
\]
Thus the full prompt is
\[
  (e_1,\ldots,e_n,e_{n+1})
  =
  \bigl((x_1,y_1),\ldots,(x_n,y_n),(x_{\mathrm{query}},0)\bigr).
\]

Assume

\begin{equation}
y_i = W^* x_i,
\end{equation}
for some unknown $W^* \in \mathbb{R}^{d_y \times d_x}$. The goal is to predict $\hat{y}_{\text{query}}$.

\paragraph{Assumptions.}
We make the following simplifying assumptions throughout this section:
\begin{enumerate}
    \item Attention is linear (unnormalized dot-product attention, i.e., no softmax).
    \item Tokens are of the form $e_i = (x_i, y_i)$, with model dimension $d_{\text{model}} = d_x + d_y$.
    \item The implicit regression weight is initialized at $W_0 = 0$.
    \item The query token attends to all training tokens and does not attend to itself; only the query token output is read.
    \item There are no normalization layers, or additional nonlinearities.
\end{enumerate}

\subsection{Gradient Descent Baseline}

Consider the squared loss
\begin{equation}
\mathcal{L}(W) = \frac{1}{2n} \sum_{i=1}^{n} \|W x_i - y_i\|^2.
\end{equation}
Starting from $W_0 = 0$, one gradient descent step with learning rate $\eta$ yields
\begin{equation}
W_1 = \frac{\eta}{n} \sum_{i=1}^{n} y_i x_i^T.
\end{equation}
The corresponding prediction is
\begin{equation}
\label{eq:gd_prediction}
\hat{y}_{\text{query}}
=
W_1 x_{\text{query}}
=
\frac{\eta}{n} \sum_{i=1}^{n} y_i (x_i^T x_{\text{query}}),
\end{equation}
which corresponds to linear kernel regression.

\subsection{Linear Self-Attention}

A linear self-attention layer with a residual connection updates token $e_j$ as
\begin{equation}
\label{eq:linear_attention}
e_j \leftarrow e_j + W_O \sum_{i=1}^{n} (q_j^T k_i) v_i,
\end{equation}
where
\[
q_j = W_Q e_j, \quad
k_i = W_K e_i, \quad
v_i = W_V e_i.
\]

Training tokens are $e_i = (x_i, y_i)$ for $i \le n$, and the query token is $e_{n+1} = (x_{\text{query}}, 0)$. The residual connection is used only to preserve the query input and is not essential to the construction.

\subsection{Main Result}

\begin{theorem}
\label{thm:linear_attention_gd}
There exist weight matrices $W_Q, W_K, W_V, W_O$ such that the linear attention update in~\eqref{eq:linear_attention}, when applied to the query token, produces (up to a fixed linear readout) the prediction in~\eqref{eq:gd_prediction}.
\end{theorem}

\paragraph{Construction.}
Let
\begin{equation}
W_Q = W_K =
\begin{bmatrix}
I_{d_x} & 0 \\
0 & 0
\end{bmatrix},
\quad
W_V =
\begin{bmatrix}
0 & 0 \\
0 & -I_{d_y}
\end{bmatrix},
\quad
W_O =
\begin{bmatrix}
0 & 0 \\
0 & \frac{\eta}{n} I_{d_y}
\end{bmatrix}.
\end{equation}

\paragraph{Proof.}
For any token $e_i = (x_i, y_i)^T$, we have
\[
q_i = k_i = (x_i, 0)^T,
\qquad
v_i = (0, -y_i)^T.
\]
Thus,
\[
q_j^T k_i = x_j^T x_i,
\]
and
\[
\sum_{i=1}^{n} (q_j^T k_i) v_i
=
\begin{bmatrix}
0 \\
- \sum_{i=1}^{n} y_i (x_i^T x_j)
\end{bmatrix}.
\]
Applying $W_O$ and evaluating the update for the query token yields
\[
e_{n+1}
=
\begin{bmatrix}
x_{\text{query}} \\
- \frac{\eta}{n} \sum_{i=1}^{n} y_i (x_i^T x_{\text{query}})
\end{bmatrix}.
\]
A final linear readout that negates the $y$-coordinates recovers the desired prediction in~\eqref{eq:gd_prediction}.
\hfill $\square$

\subsection{Limitations}
\label{sec:limitations}

The construction in Theorem~\ref{thm:linear_attention_gd} is highly idealized and should not be viewed as an explanation of practical ICL:
\begin{enumerate}
    \item Softmax attention introduces normalization and nonlinearity absent from linear attention.
    \item Linear attention cannot selectively focus on relevant examples or ignore outliers. ICL often requires selective attention to appropriate demonstrations.
    \item The construction  excludes architectural components standard in modern transformers.

    \item The construction only implements one gradient descent step. Extending to multi-layer transformers requires different techniques and does not compositionally build deeper gradient descent procedures.
    
\end{enumerate}

\section{Additional Results on Final Normalization}
\label{app:final_norme}

\subsection{Boundary values with RMSNorm}

\begin{figure}[!ht]
\centering
\includegraphics[width=0.5\textwidth]{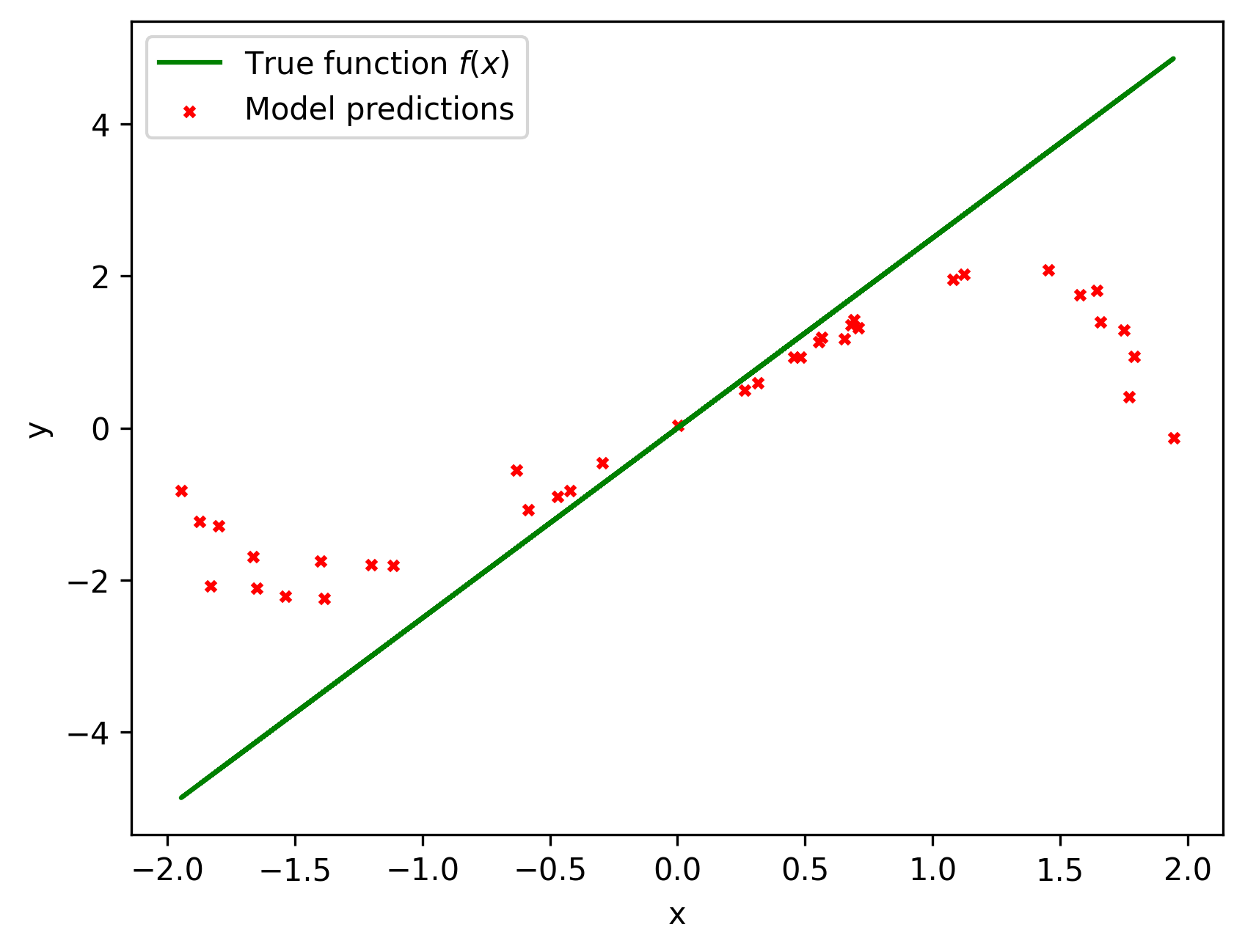}
\caption{Boundary Values for 12L8AH Transformer with RMSNorm}
\label{rmsnorm}

\end{figure}

%\newpage

\subsection{Changing the Final Normalization Scale}
\label{app:final_norm_scale}
%\paragraph{Changing the final-normalization scale.}
A remaining possibility is that the final normalized readout range is too small. For architectures with a final normalization layer, the readout bound depends on the scale and shift parameters of that layer.
Increasing the normalization scale enlarges the set of normalized representations accessible to the decoder. We therefore test whether increasing this scale can mitigate the failure. In the LayerNorm models used here, this corresponds to manipulating the final LayerNorm scale parameter $\gamma$: we consider post-hoc scaling of $\gamma$, training with $\gamma$ frozen at a larger value, and large-$\gamma$ initialization with learning.

\begin{table}[!ht]
  \centering
  \small
  \begin{tabular}{lcccccc}
    \toprule
    & $\sigma{=}1$ & $\sigma{=}2$ & $\sigma{=}3$ & $\sigma{=}5$
    & $\sigma{=}7$ & $\sigma{=}10$ \\
    \midrule
    Baseline                         & $9{\times}10^{-6}$ & $0.04$ & $0.62$ & $4.29$ & $10.36$ & $27.06$ \\
    Post-hoc $1.5\times\gamma$       & $0.18$ & $0.41$ & $0.41$ & $2.54$ & $6.35$ & $22.78$ \\
    Post-hoc $2\times\gamma$         & $0.73$ & $1.88$ & $1.50$ & $2.19$ & $4.45$ & $17.18$ \\
    Post-hoc $5\times\gamma$         & $11.2$ & $30.9$ & $31.0$ & $26.9$ & $22.8$ & $16.3$  \\
    Frozen $\gamma{=}2$ (retrained)  & $9{\times}10^{-6}$ & $0.05$ & $0.85$ & $5.76$ & $13.22$ & $38.68$ \\
    Large $\gamma$ init (learned)    & $3{\times}10^{-5}$ & $0.05$ & $0.93$ & $6.45$ & $13.58$ & $37.79$ \\
    \bottomrule
  \end{tabular}
  \caption{Effect of changing the final-normalization scale on MSE for degree-1 polynomials. In these experiments, we manipulate the final LayerNorm scale parameter $\gamma$. Post-hoc scaling changes the accessible output range but damages in-distribution accuracy. Retraining with frozen or large-initialized $\gamma$ does not restore OOD generalization, indicating that the failure is not solved by simply increasing the final-normalization scale.}
  \label{tab:gamma}
\end{table}

Table~\ref{tab:gamma} shows that increasing the final-normalization scale does not restore distribution-independent ICL. Post-hoc scaling can shift the effective output range, but it damages in-distribution accuracy. Retraining with frozen or large-initialized $\gamma$ also fails to improve OOD performance. Thus, although final normalization controls the hard readout bound, simply increasing its scale does not make the model behave like a scale-equivariant regression algorithm.

\newpage
\section{Proof of Theorem~\ref{thm:boundedness}}
\label{app:proofTh1}

We prove a boundedness result for standard Transformer architectures whose predictions are decoded from a final normalized query representation. This setting includes both conventional Transformer models and the architectures described in Section~\ref{sec:normalizationA} that employ a final LayerNorm or RMSNorm. Notably, the proof does not depend on the behavior of internal normalization layers; it relies solely on the assumption that the representation fed to the decoder lies within a compact set.

Let $X_L(x) \in \mathbb{R}^{n \times d}$ denote the final residual-stream representation produced by the transformer on an input sequence with query $x$, where $n$ is the sequence length and $d$ is the model dimension. Let $q$ denote the query-token position. The scalar prediction has the form
\[
    \hat{g}(x)
    =
    \text{Dec}\!\left(
        \mathrm{Norm}_{\mathrm{final}}(X_L(x)_q)
      \right),
\]
where $\text{Dec}:\mathbb{R}^d \to \mathbb{R}$ is the decoder and $\mathrm{Norm}_{\mathrm{final}}$ is either LayerNorm or RMSNorm.

We first show that the final normalized representation is uniformly bounded.

\paragraph{Restatement of Lemma~\ref{lemma:compact}.}
For any hidden representation $z\in\mathbb{R}^d$, LayerNorm with parameters $\gamma,\beta\in\mathbb{R}^d$,
\[
\mathrm{LN}(z)
=
\gamma \odot
\frac{z-\mu(z)\mathbf{1}}
{\sqrt{\sigma^2(z)+\epsilon}}
+\beta,
\]
satisfies $\mathrm{LN}(z)\in K_{\mathrm{LN}}$ for all $z\in\mathbb{R}^d$, where
\[
\mathcal{K}_{\mathrm{LN}}
=
\left\{
y\in\mathbb{R}^d:
\|y\|_2 \leq \|\gamma\|_\infty\sqrt{d}+\|\beta\|_2
\right\}.
\]

\begin{proof}
LayerNorm is defined by
\[
\mathrm{LN}(z)
=
\gamma \odot
\frac{z-\mu(z)\mathbf{1}}{\sqrt{\sigma^2(z)+\varepsilon}}
+
\beta,
\]
where
\[
\mu(z)=\frac{1}{d}\sum_{i=1}^d z_i,
\qquad
\sigma^2(z)=\frac{1}{d}\sum_{i=1}^d (z_i-\mu(z))^2,
\qquad
\varepsilon>0.
\]
Define
\[
    \widetilde{z}
    =
    \frac{z-\mu(z)\mathbf{1}}{\sqrt{\sigma^2(z)+\varepsilon}}.
\]
Then
\[
\|\widetilde{z}\|_2^2
=
\frac{\sum_{i=1}^d (z_i-\mu(z))^2}
     {\sigma^2(z)+\varepsilon}
=
\frac{d\,\sigma^2(z)}
     {\sigma^2(z)+\varepsilon}
\leq d.
\]
Therefore $\|\widetilde{z}\|_2 \leq \sqrt{d}$. Hence
\[
\|\mathrm{LN}(z)\|_2
=
\|\gamma \odot \widetilde{z}+\beta\|_2
\leq
\|\gamma\|_\infty \|\widetilde{z}\|_2 + \|\beta\|_2
\leq
\|\gamma\|_\infty\sqrt{d}+\|\beta\|_2.
\]
The set $\mathcal{K}_{\mathrm{LN}}$ is closed and bounded in finite dimension, hence compact.
\end{proof}

We now prove Theorem~\ref{thm:boundedness}. Suppose first that $\mathrm{Norm}_{\mathrm{final}}=\mathrm{LN}$. By Lemma~\ref{lemma:compact},
\[
    \mathrm{LN}(X_L(x)_q) \in \mathcal{K}_{\mathrm{LN}}
    \qquad
    \text{for all inputs } x.
\]
Since Dec is continuous and $\mathcal{K}_{\mathrm{LN}}$ is compact, $\text{Dec}(\mathcal{K}_{\mathrm{LN}})$ is compact in $\mathbb{R}$. Hence it is bounded. Therefore there exists $B>0$ such that
\[
    |\text{Dec}(y)| \leq B
    \qquad
    \text{for all } y \in \mathcal{K}_{\mathrm{LN}}.
\]
Taking
\[
    y=\mathrm{LN}(X_L(x)_q)
\]
gives
\[
    |\hat{g}(x)| \leq B
    \qquad
    \text{for all inputs } x.
\]
This proves the theorem for final LayerNorm.

\subsection*{Extension to RMSNorm}

The same argument applies to RMSNorm. RMSNorm is defined by
\[
\mathrm{RMSNorm}(z)
=
\gamma \odot
\frac{z}{\sqrt{\frac{1}{d}\sum_{i=1}^d z_i^2+\varepsilon}},
\]
where $\gamma\in\mathbb{R}^d$ and $\varepsilon>0$. Define
\[
    \widetilde{z}_{\mathrm{RMS}}
    =
    \frac{z}{\sqrt{\frac{1}{d}\sum_{i=1}^d z_i^2+\varepsilon}}.
\]
Then
\[
\|\widetilde{z}_{\mathrm{RMS}}\|_2^2
=
\frac{\sum_{i=1}^d z_i^2}
     {\frac{1}{d}\sum_{i=1}^d z_i^2+\varepsilon}
\leq d,
\]
so
\[
\|\mathrm{RMSNorm}(z)\|_2
=
\|\gamma \odot \widetilde{z}_{\mathrm{RMS}}\|_2
\leq
\|\gamma\|_\infty\sqrt{d}.
\]
Thus the image of RMSNorm is contained in the compact set
\[
    \mathcal{K}_{\mathrm{RMS}}
    =
    \left\{
        y\in\mathbb{R}^d:
        \|y\|_2 \leq \|\gamma\|_\infty\sqrt{d}
    \right\}.
\]
Since the decoder $Dec$ is continuous, $Dec(\mathcal{K}_{\mathrm{RMS}})$ is compact and therefore bounded. Hence there exists $B>0$ such that
\[
    |\hat{g}(x)| \leq B
    \qquad
    \text{for all inputs } x.
\]
This proves Theorem~\ref{thm:boundedness} for final RMSNorm as well.
\qed

% -----------------------------------------------------------------
\section{Proof of Corollary~\ref{cor:icl2}}
\label{app:proofNOICL2}

\begin{corollary}
\label{cor:icl2}
Assume the conditions of Theorem~\ref{thm:boundedness}. Then a transformer whose scalar prediction is decoded from a final normalized representation cannot achieve $\mathrm{ICL}_2$ for any non-constant polynomial target $f:\mathbb{R}^d\to\mathbb{R}$ over the full domain $\mathbb{R}^d$.
\end{corollary}

\begin{proof}
By Theorem~\ref{thm:boundedness}, there exists a constant $B>0$ such that the model prediction satisfies
\[
    |\hat{g}(x)| \leq B
    \qquad
    \text{for all } x.
\]
Let $f:\mathbb{R}^d\to\mathbb{R}$ be a non-constant polynomial. Then $f$ is unbounded on $\mathbb{R}^d$. Indeed, since $f$ is non-constant, there exists a direction $v\in\mathbb{R}^d$ and a point $x_0\in\mathbb{R}^d$ such that the univariate restriction
\[
    p(t) = f(x_0 + tv)
\]
is a non-constant polynomial in $t$. Every non-constant univariate polynomial is unbounded in magnitude as $|t|\to\infty$. Hence, for any $B>0$ and any target accuracy $\varepsilon>0$, there exists $x^\star\in\mathbb{R}^d$ such that
\[
    |f(x^\star)| > B+\varepsilon.
\]
For this input,
\[
|\hat{g}(x^\star)-f(x^\star)|
\geq
|f(x^\star)|-|\hat{g}(x^\star)|
>
\varepsilon.
\]
Therefore the model cannot uniformly approximate $f$ over
$\mathbb{R}^d$ to arbitrary accuracy. In particular, it cannot satisfy $\mathrm{ICL}_2$ for non-constant polynomial functions over an unbounded domain.
\end{proof}

\section{Attention Collapse Under Distribution Shift}
\label{app:attention_collapse}

\subsection{Post-Hoc and Training-Time Attention Smoothing}
\label{app:attention_smoothing}

The attention diagnostics in Section~\ref{sec:attention-bottleneck} show that attention becomes increasingly concentrated under distribution shift. We next ask whether reducing this concentration can mitigate the OOD failure. We test this in two complementary ways: by smoothing the attention distribution of an already trained model at inference time, and by training a model from scratch
with temperature-scaled softmax attention.

\paragraph{Post-hoc attention smoothing.}
We first intervene on each standard trained model at inference time by temperature-scaling the attention scores,
\[
    a_j(T)
    =
    \frac{\exp(z_j/T)}
    {\sum_k \exp(z_k/T)}.
\]
We evaluate
\[
    T \in \{1,2,5,20,10^6\},
\]
where $T=1$ recovers the original attention distribution and increasing $T$ progressively flattens it. All learned parameters are kept fixed.

We apply this intervention under the coefficient-scale shift for polynomial degrees $1$--$6$. The results are consistent across degrees (Table~\ref{tab:attention_smoothing}). Increasing the temperature further does not improve OOD generalization; instead, stronger smoothing progressively degrades both in-distribution and OOD performance.

\begin{table}[!ht]
\centering
\scriptsize
\setlength{\tabcolsep}{3.2pt}
\begin{tabular}{clcccccccccc}
\toprule
Degree & Temperature
& $\sigma=1$ & $2$ & $3$ & $4$ & $5$
& $6$ & $7$ & $8$ & $9$ & $10$ \\
\midrule

\multirow{5}{*}{1}
& $T=2$
& $2{\times}10^{-4}$ & $0.06$ & $0.64$ & $2.05$ & $4.20$
& $6.68$ & $11.85$ & $15.93$ & $21.58$ & $29.99$ \\
& $T=5$
& $0.01$ & $0.12$ & $0.87$ & $2.51$ & $4.90$
& $7.60$ & $13.09$ & $17.42$ & $23.29$ & $31.97$ \\
& $T=20$
& $0.15$ & $0.74$ & $2.10$ & $4.36$ & $7.33$
& $10.58$ & $16.77$ & $21.68$ & $28.05$ & $37.32$ \\
& $T=10^6$
& $0.27$ & $1.19$ & $2.93$ & $5.58$ & $8.93$
& $12.42$ & $19.15$ & $24.32$ & $30.97$ & $40.75$ \\
\midrule

\multirow{5}{*}{2}
& $T=2$
& $5{\times}10^{-4}$ & $0.07$ & $0.72$ & $2.50$ & $4.94$
& $8.77$ & $14.47$ & $18.95$ & $27.13$ & $34.62$ \\
& $T=5$
& $0.01$ & $0.17$ & $1.03$ & $3.03$ & $5.68$
& $9.74$ & $15.71$ & $20.35$ & $28.81$ & $36.47$ \\
& $T=20$
& $0.12$ & $0.70$ & $2.03$ & $4.45$ & $7.56$
& $11.97$ & $18.37$ & $23.36$ & $32.38$ & $40.26$ \\
& $T=10^6$
& $0.31$ & $1.28$ & $2.94$ & $5.67$ & $8.93$
& $13.48$ & $19.92$ & $24.87$ & $33.78$ & $41.61$ \\
\midrule

\multirow{5}{*}{3}
& $T=2$
& $4{\times}10^{-4}$ & $0.04$ & $0.40$ & $1.54$ & $3.43$
& $6.33$ & $10.80$ & $15.29$ & $22.70$ & $28.74$ \\
& $T=5$
& $6{\times}10^{-3}$ & $0.09$ & $0.62$ & $2.00$ & $4.18$
& $7.40$ & $12.27$ & $17.11$ & $24.90$ & $31.24$ \\
& $T=20$
& $0.07$ & $0.54$ & $1.73$ & $3.92$ & $6.95$
& $10.93$ & $16.68$ & $22.38$ & $30.95$ & $37.80$ \\
& $T=10^6$
& $0.34$ & $1.23$ & $2.88$ & $5.60$ & $9.19$
& $13.62$ & $19.89$ & $26.10$ & $35.30$ & $42.29$ \\
\midrule

\multirow{5}{*}{4}
& $T=2$
& $2{\times}10^{-3}$ & $0.05$ & $0.49$ & $1.65$ & $3.79$
& $7.04$ & $11.48$ & $15.64$ & $22.82$ & $31.39$ \\
& $T=5$
& $0.06$ & $0.26$ & $0.96$ & $2.42$ & $4.95$
& $8.51$ & $13.43$ & $17.99$ & $25.63$ & $34.55$ \\
& $T=20$
& $0.29$ & $1.06$ & $2.65$ & $5.01$ & $8.63$
& $12.84$ & $18.54$ & $23.92$ & $32.69$ & $42.40$ \\
& $T=10^6$
& $0.34$ & $1.29$ & $3.15$ & $5.92$ & $9.90$
& $14.37$ & $20.45$ & $26.10$ & $35.12$ & $45.03$ \\
\midrule

\multirow{5}{*}{5}
& $T=2$
& $2{\times}10^{-3}$ & $0.03$ & $0.41$ & $1.50$ & $3.99$
& $7.68$ & $11.97$ & $16.95$ & $25.12$ & $35.21$ \\
& $T=5$
& $0.02$ & $0.10$ & $0.68$ & $2.03$ & $4.95$
& $9.01$ & $13.75$ & $19.05$ & $27.55$ & $37.97$ \\
& $T=20$
& $0.27$ & $1.01$ & $2.66$ & $4.97$ & $9.31$
& $14.07$ & $19.83$ & $25.74$ & $35.05$ & $46.22$ \\
& $T=10^6$
& $0.49$ & $1.58$ & $3.79$ & $6.64$ & $11.77$
& $17.29$ & $23.37$ & $29.97$ & $39.86$ & $51.16$ \\
\midrule

\multirow{5}{*}{6}
& $T=2$
& $3{\times}10^{-3}$ & $0.04$ & $0.48$ & $1.82$ & $4.62$
& $8.75$ & $13.71$ & $19.96$ & $28.81$ & $37.85$ \\
& $T=5$
& $0.02$ & $0.12$ & $0.76$ & $2.38$ & $5.45$
& $9.77$ & $15.05$ & $21.48$ & $30.46$ & $39.67$ \\
& $T=20$
& $0.15$ & $0.83$ & $2.52$ & $5.22$ & $9.26$
& $14.36$ & $20.42$ & $27.60$ & $37.10$ & $46.70$ \\
& $T=10^6$
& $0.43$ & $1.50$ & $3.64$ & $6.70$ & $10.97$
& $16.21$ & $22.27$ & $29.65$ & $38.93$ & $48.60$ \\
\bottomrule
\end{tabular}
\caption{
Inference-time attention smoothing across polynomial degrees. Attention logits are divided by a temperature $T$. The case $T=1$ is the original model (see Table~\ref{table:3a}).
}
\label{tab:attention_smoothing}
\end{table}

%\newpage

\paragraph{Training with temperature-scaled softmax.}
A possible limitation of the post-hoc intervention is that the model was trained with the original softmax attention and therefore has no opportunity to adapt its representations to the modified attention distribution. We perform a complementary experiment in which the model is trained from scratch with temperature-scaled softmax attention. In this setting, the modified attention rule is present throughout optimization, allowing the remaining model parameters to adapt to the flatter attention distribution. Despite this adaptation, the retrained model still exhibits strong degradation under distribution shift (Table \ref{table:LF}). 

Taken together, these experiments show that simply flattening attention is not sufficient to recover distribution-independent in-context regression: neither post-hoc smoothing nor training the model to operate with a flatter attention distribution restores $\mathrm{ICL}_2$.

\subsection{Attention-mechanism variants}
\label{app:attention_variants}

The analysis in Section~\ref{sec:attention-bottleneck} focuses on standard dot-product softmax attention. To test whether the observed failure is an artifact of this specific attention normalization, we evaluate several attention variants: linear attention, temperature-scaled softmax, sparsemax, entmax, and CosFormer-style attention.

All variants are trained from scratch using the same protocol as the main models. This distinction is important: the attention mechanism is not modified post hoc after training, but is present throughout optimization, allowing the model weights to adapt to the chosen attention rule. We then evaluate the resulting models under the same conditions used in the main experiments.

Across these variants, we observe the same qualitative pattern as for the standard transformer. The models achieve strong in-distribution ICL on the training distribution, but still degrade sharply under distribution shift. Temperature scaling changes the sharpness of the attention distribution, sparsemax and entmax change its sparsity and support, and CosFormer changes the attention parameterization more substantially. Nevertheless, none of these modifications restores the scale-equivariant behavior expected of least-squares or polynomial regression.

\begin{table}[!ht]
\centering
\resizebox{\linewidth}{!}{
\begin{tabular}{l|l|l|l|l|l|l|l|l|l|l}
 \hline
 %\rowcolor{orange!30}
 models \ $\backslash$ \ $\sigma$ & 1 & 2 & 3 & 4 & 5 & 6 & 7 & 8 & 9 & 10 \\ 
 \hline
%  SSA   & $4\times10^{-5}$& $3\times10^{-4}$& $10^{-3}$& $0.02$& $0.02$ & $0.15$& $1.24$& $1.04$& $2.74$& $8.50$ \\

 Softmax   & $9 \times 10^{-6}$& 0.04 & 0.62& 1.95& 4.29& 7.14& 10.36& 15.79& 24.48& 27.06 \\ 
 
 Linear Attention & $9 \times 10^{-6}$ &0.03 & 0.53 & 1.66 & 3.47 & 6.26 & 11.46 & 15.30 & 23.13 &29.71 \\
 Sparsemax & $1 \times 10^{-4}$ & 0.06 & 0.68 & 2.24 & 4.57 & 7.62 & 13.43 & 18.12 & 24.76 & 32.93 \\
 1.5-Entmax & $4\times10^{-6}$& 0.07& 0.76 & 2.17 & 4.29 & 6.78 & 11.66 & 15.61& 21.22& 28.06 \\
Temperature Softmax  & $7\times 10^{-6}$ & 0.07 & 0.83 &  2.45 & 4.81 & 7.54 &  12.89 & 17.05 & 22.91 & 30.41
\\

CosFormer &  $7\times10^{-6}$ & 0.05 & 0.77 & 2.58 & 5.56 & 9.21 & 16.14 & 21.19 & 28.53 & 37.50 \\ 
 \hline
\end{tabular}
}
\caption{MSE for degree-1 regression under distribution shift with different attention mechanisms. Each model is trained from scratch using its specified attention rule throughout training and evaluation. Training inputs and coefficients are sampled from $\mathcal{U}(-1,1)$. At test time, inputs remain sampled from $\mathcal{U}(-1,1)$, while coefficients are sampled from $\mathcal{U}(-\sigma,\sigma)$.}
\label{table:LF}
\end{table}

\section{Asymptotic Analysis of a Single Attention Layer}
\label{sec:mathproofs}

The following lemma isolates a bottleneck effect of attention in a minimal setting.

\begin{lemma}[Asymptotic form of one attention layer]
\label{lemma5}
Consider a single attention layer with $H$ softmax heads acting on scalar tokens $(x_1,\ldots,x_p,x)$, $p\ge 1$, where $x_1,\ldots,x_p$ are the context tokens (inputs and labels) and $x$ is the query input. Assume that (i) every token is embedded by the same bias-free linear map $t\mapsto tW$, $W\in\mathbb{R}^{1\times d}$, with no positional encoding; (ii) the query, key, value and output projections are bias-free; (iii) the query position attends to all context tokens and to itself; and (iv) no normalization is applied. Then, for fixed context tokens,
\[
    \mathrm{Attn}_p(x)=x\,a_p+b_\pm+o(1)\qquad (x\to\pm\infty),
\]
where $a_p\in\mathbb{R}^{1\times d}$ depends only on the model parameters and on $p$, not on the values $x_1,\ldots,x_p$, whereas $b_\pm\in\mathbb{R}^{1\times d}$ depend on the context.
\end{lemma}

The lemma shows that, in a large-query regime, the coefficient of the leading term is fixed by the parameters and the context length; the in-context examples enter only at order one.

\begin{proof}
For head $h$, let $Q^h,K^h\in\mathbb{R}^{d\times d_k}$, $V^h\in\mathbb{R}^{d\times d_v}$ and $O^h\in\mathbb{R}^{d_v\times d}$ be the query, key, value and output projections, and define
\[
\alpha_h := \frac{1}{\sqrt{d_k}}\,WQ^hK^{h\top}W^\top\in\mathbb{R},
\qquad
\zeta_h := WV^hO^h\in\mathbb{R}^{1\times d}.
\]
Since token $t$ is embedded as $tW$, the scores at the query position are $\alpha_h x x_j$ for context token $j$ and $\alpha_h x^2$ for the query token itself. With
\[
Z_h(x)=e^{\alpha_h x^2}+\sum_{k=1}^p e^{\alpha_h x x_k},
\qquad
w_{\mathrm{self}}(x)=\frac{e^{\alpha_h x^2}}{Z_h(x)},
\qquad
w_j(x)=\frac{e^{\alpha_h x x_j}}{Z_h(x)},
\]
the output of head $h$ at the query position is
\[
\mathrm{Attn}_p^h(x)=\Bigl(x\,w_{\mathrm{self}}(x)+\sum_{j=1}^p x_j\,w_j(x)\Bigr)\zeta_h,
\qquad
\mathrm{Attn}_p(x)=\sum_{h=1}^H \mathrm{Attn}_p^h(x).
\]

\paragraph{Case 1: $\alpha_h>0$.}
Let $R=2\max_j|x_j|$. For $|x|>R$ we have $|x_j/x|\le 1/2$, hence $\alpha_h(x^2-xx_j)\ge \alpha_h x^2/2$ for all $j$. Since $Z_h(x)\ge e^{\alpha_h x^2}$,
\[
w_j(x)\le e^{-\alpha_h(x^2-xx_j)}\le e^{-\alpha_h x^2/2},
\qquad
1-w_{\mathrm{self}}(x)=\sum_{j=1}^p w_j(x)\le p\,e^{-\alpha_h x^2/2}.
\]
Therefore $x\bigl(w_{\mathrm{self}}(x)-1\bigr)\to 0$ and $\sum_j x_j w_j(x)\to 0$, so
\[
\mathrm{Attn}_p^h(x)=x\,\zeta_h+o(1)\qquad (x\to\pm\infty).
\]

\paragraph{Case 2: $\alpha_h<0$.}
Let $m_+=\min_j x_j$, $m_-=\max_j x_j$, and $I_+=\{j: x_j=m_+\}$. Consider $x\to+\infty$. Since $Z_h(x)\ge e^{\alpha_h x m_+}$, every context token $j\notin I_+$ satisfies
\[
w_j(x)\le e^{\alpha_h x(x_j-m_+)}\le e^{-|\alpha_h|\delta_+ x},
\qquad
\delta_+=\min_{j\notin I_+}(x_j-m_+)>0,
\]
and, for $x\ge 2|m_+|$, the query token satisfies
\[
w_{\mathrm{self}}(x)\le e^{\alpha_h x(x-m_+)}\le e^{-|\alpha_h| x^2/2}.
\]
Hence $x\,w_{\mathrm{self}}(x)\to0$, $\sum_{j\notin I_+}x_jw_j(x)\to0$, and $\sum_{j\in I_+}w_j(x)\to1$. Since all tokens in $I_+$ have the value $m_+$,
\[
\mathrm{Attn}_p^h(x)=m_+\zeta_h+o(1)\qquad (x\to+\infty).
\]
The case $x\to-\infty$ is symmetric, with $m_-$ and $I_-=\{j: x_j=m_-\}$ in place of $m_+$ and $I_+$, and gives $\mathrm{Attn}_p^h(x)=m_-\zeta_h+o(1)$. (If all context tokens are equal, the bounds on $j\notin I_\pm$ are vacuous.)

\paragraph{Case 3: $\alpha_h=0$.}
All scores are equal, so attention is uniform and
\[
\mathrm{Attn}_p^h(x)=\frac{1}{p+1}\Bigl(x+\sum_{j=1}^p x_j\Bigr)\zeta_h.
\]

\paragraph{Conclusion.}
Summing over the finitely many heads gives $\mathrm{Attn}_p(x)=x\,a_p+b_\pm+o(1)$ as $x\to\pm\infty$, with
\[
a_p=\sum_{h:\alpha_h>0}\zeta_h+\sum_{h:\alpha_h=0}\frac{\zeta_h}{p+1},
\qquad
b_\pm=m_\pm\sum_{h:\alpha_h<0}\zeta_h+\Bigl(\sum_{j=1}^p x_j\Bigr)\sum_{h:\alpha_h=0}\frac{\zeta_h}{p+1}.
\]
The coefficient $a_p$ does not arise from solving a regression problem over the in-context examples; the context enters only through $b_\pm$. Heads with $\alpha_h>0$ concentrate on the query token itself, and heads with $\alpha_h<0$ on the extremal context token(s).
\end{proof}

\section{Architectural Interventions with Additional Inductive Biases}
\label{app:architectural_interventions}

We test in this section two other architectural interventions with stronger bias. The first one is a permutation-invariant set regressor that uses neither attention nor final normalization. The second one is a hybrid transformer-ridge in which a ridge-regression fitting step is implemented into the architecture.

\paragraph{Permutation-invariant set regressor.}
The set regressor summarizes the preceding context pairs using a shared encoder and mean pooling. For the context available at position $i$, it computes
\[
    r_i
    =
    \frac{1}{|\mathcal{C}_i|}
    \sum_{(x_j,y_j)\in\mathcal{C}_i}
    \phi(x_j,y_j),
    \qquad
    \widehat{y}_i=\rho(x_i,r_i),
\]
where $\mathcal{C}_i=\{(x_j,y_j):j<i\}$ and $r_i=0$ when $\mathcal{C}_i$ is empty. Both $\phi$ and $\rho$ are four-layer MLPs with hidden width 256. The architecture uses no attention, positional encoding, feature normalization, or final normalized readout.
It therefore tests whether permutation-invariant context aggregation alone is sufficient to obtain distribution-independent regression.

\paragraph{Transformer-ridge hybrid.}
The hybrid separates feature learning from fitting the context labels. A 12-layer, 8-head Transformer with hidden dimension 256 receives only the sequence of inputs and produces a 16-dimensional feature vector for each position. We add a constant feature to obtain $\psi_j\in\mathbb{R}^{17}$, which depends only on $x_1,\dots,x_j$. At query position $i$, we stack the features of the preceding positions as the rows of $\Psi_{<i}\in\mathbb{R}^{(i-1)\times17}$ and define $y_{<i}=(y_1,\dots,y_{i-1})^\top$. We then fit ridge regression in closed form,
\[
    \widehat{w}_i
    =
    \left(
        \Psi_{<i}^{\top}\Psi_{<i}+\lambda I_{17}
    \right)^{-1}
    \Psi_{<i}^{\top}y_{<i},
    \qquad
    \widehat{y}_i=\psi_i^{\top}\widehat{w}_i,
\]
with $\lambda=10^{-3}$. The feature extractor never receives the labels, which enter only through the closed-form ridge solve. Consequently, multiplying all context labels by $\alpha$ multiplies every prediction by $\alpha$ by construction.

\begin{table}[!ht]
\centering
\small
\setlength{\tabcolsep}{3.5pt}
\resizebox{\textwidth}{!}{%
\begin{tabular}{lrrrrrrrrrr}
\toprule
Model
& $\sigma=1$
& $\sigma=2$
& $\sigma=3$
& $\sigma=4$
& $\sigma=5$
& $\sigma=6$
& $\sigma=7$
& $\sigma=8$
& $\sigma=9$
& $\sigma=10$ \\
\midrule
Set regressor (CNP)
& $7{\times}10^{-5}$
& $0.26$
& $2.27$
& $4.46$
& $7.77$
& $12.42$
& $19.79$
& $24.99$
& $34.90$
& $41.92$ \\
Transformer--ridge hybrid
& $7{\times}10^{-4}$
& $3{\times}10^{-3}$
& $7{\times}10^{-3}$
& $0.01$
& $0.02$
& $0.02$
& $0.04$
& $0.05$
& $0.06$
& $0.08$ \\
\bottomrule
\end{tabular}%
}
\caption{MSE under coefficient scaling for the permutation-invariant set regressor and Transformer-ridge hybrid. Both models are trained on degree-1 functions with coefficients sampled from $\mathcal{U}(-1,1)$ and evaluated with coefficients sampled from $\mathcal{U}(-\sigma,\sigma)$.}
\label{tab:architectural_interventions}
\end{table}

Table \ref{tab:architectural_interventions} shows that removing softmax attention and the final normalized readout is not sufficient to prevent degradation under coefficient scaling. The Transformer-ridge hybrid remains accurate under this shift because its architecture explicitly incorporates the appropriate fitting rule. The Transformer learns the feature representation, while ridge is performed by construction. This shows that a suitably chosen inductive bias can enforce robustness
along the shift covered by the built-in estimator. In many real applications of ICL, however, the appropriate fitting algorithm is not known in advance, and the model is expected to infer from the examples in context how to make a prediction.

%%%%%%%%%%%%%%%%%%%%%%%%%%%%%%%%%%%%%%%%%%%%%%%%%%%%%%%%%%%%

%\newpage
%\input{checklist.tex}

\end{document}